\pdfoutput=1

\documentclass[11pt]{article}

\usepackage[]{acl}

\usepackage{times}
\usepackage{latexsym}

\usepackage[T1]{fontenc}

\usepackage[utf8]{inputenc}

\usepackage{microtype}
\usepackage{url}
\newcommand\R{\mathbb{R}}
\usepackage{graphicx}
\usepackage{multirow}
\usepackage{booktabs}
\usepackage{amsmath,amssymb}
\graphicspath{ {./pictures/} }
\usepackage{array}
\newcolumntype{P}[1]{>{\centering\arraybackslash}p{#1}}
\usepackage{url}
\usepackage{xcolor}
\usepackage{float}
\usepackage{caption,subcaption}
\usepackage{courier}
\usepackage{soul}
\usepackage{colortbl}
\definecolor{r}{RGB}{240,160,180}
\definecolor{y}{RGB}{240,225,140}
\definecolor{b}{RGB}{130,220,240}
\definecolor{p}{RGB}{220,190,230}

\title{Balancing Multi-Domain Corpora Learning for Open-Domain Response Generation}

\newcommand\blfootnote[1]{%
  \begingroup
  \renewcommand\thefootnote{}\footnote{#1}%
  \addtocounter{footnote}{-1}%
  \endgroup
}

\author{Yujie Xing$^{1}$,\ Jinglun Cai$^{2*}$,\ Nils Barlaug$^{1}$,\ Peng Liu$^{1}$,\ Jon Atle Gulla$^{1}$ \\
  $^{1}$Department of Computer Science, Norwegian University of Science and Technology\\
  $^{2}$Amazon AWS AI\\
  \texttt{\{yujie.xing,nils.barlaug,peng.liu,jon.atle.gulla\}@ntnu.no}\\\texttt{cjinglun@amazon.com}}

\begin{document}
\maketitle
\begin{abstract}
Open-domain\blfootnote{* This work was done prior to the author joining Amazon.} conversational systems are assumed to generate equally good responses on multiple domains. Previous work achieved good performance on the single corpus, but training and evaluating on multiple corpora from different domains are less studied. This paper explores methods of generating relevant responses for each of multiple multi-domain corpora. We first examine interleaved learning which intermingles multiple corpora as the baseline. We then investigate two multi-domain learning methods, labeled learning and multi-task labeled learning, which encode each corpus through a unique corpus embedding. Furthermore, we propose Domain-specific Frequency (DF), a novel word-level importance weight that measures the relative importance of a word for a specific corpus compared to other corpora. Based on DF, we propose weighted learning, a method that integrates DF to the loss function. We also adopt DF as a new evaluation metric. Extensive experiments show that our methods gain significant improvements on both automatic and human evaluation. We share our code and data for reproducibility.\footnote{\url{https://github.com/yujie-xing/Balancing_Multi_Domain_Corpus_Learning_for_Open_Domain_Response_Generation}}
\end{abstract}

\section{Introduction}

Recent work has achieved improvements in general performance for open-domain response generation \citep{le,iulian2,li-persona,mix}. However, most studies are restricted to single-corpus training and evaluating, while there lacks studies for training and evaluating with multiple corpora from different domains. Single-corpus training has intrinsic limitations. For example, a corpus of everyday chats, e.g., the PersonaChat corpus \citep{convai2}, does not cover technical topics discussed in Ubuntu chatlogs \citep{ubuntu}. A conversational system that learns only from PersonaChat or from multiple corpora without an appropriate technique is not likely to generate relevant responses for certain topics (see Table \ref{table:intro0}). Therefore, it is necessary for an open-domain conversational system to learn from multiple corpora, and to learn with good techniques.

\begin{table} \centering
{\footnotesize
    \begin{tabular}{p{1.9 cm} p{0.30\textwidth}}
    \toprule
        Context & What are you going to do on the remote system exactly?\\
    \hline\hline
        PersonaChat & I am going to be a pilot. I am going to fly planes.\\
        4 corpora (concatenated) & I am going to go to the beach.\\
    \toprule
    \end{tabular}
}
\caption{Irrelevant responses generated from fine-tuned GPT-2. The GPT-2 model is fine-tuned respectively on PersonaChat / concatenated 4 corpora (OpenSubtitles, Twitter, Ubuntu, PersonaChat)}\label{table:intro0}

\begin{tabular}{c}
\quad\\
\end{tabular}
{\footnotesize
\begin{tabular}{l|P{0.6cm}P{0.6cm}P{0.6cm}P{1.2cm}}
            & \multicolumn{4}{c}{Test set}\\
    Fine-tune corpus & OSDB & Twitter & Ubuntu & PersonaChat\\
    \hline
        PersonaChat & 478.8 & 159.6 & 264.7 & 19.6\\
    \hline
        \multirow{2}{2cm}{4 corpora (concatenated)} & \multirow{2}{*}{392.8} & \multirow{2}{*}{110.7} & \multirow{2}{*}{199.2} & \multirow{2}{*}{19.0}\\
        & & & & \\
\end{tabular}
}

\caption{Imbalanced perplexity performance of fine-tuned GPT-2. The GPT-2 model is fine-tuned on PersonaChat / concatenated 4 corpora (OpenSubtitles, Twitter, Ubuntu, PersonaChat)}\label{table:intro}
\end{table}

Furthermore, the case of using a single small-scale open-domain corpus has apparent weaknesses. A common way of dealing with a small-scale corpus is through fine-tuning \citep{li-persona,pure-fine-tune,mix-fine-tune}. Fine-tuning on a single corpus tends to make the model overfit on that specific corpus while performing worse on other corpora. Table \ref{table:intro} shows the result of a GPT-2 model gaining good performance on PersonaChat while performing poorly on other corpora.

This paper explores how to train and evaluate on multiple corpora from different domains for the open-domain response generation task. We propose several methods to make a model generate relevant responses for each of the multiple corpora.

Since simply training multiple corpora one by one does not solve the imbalanced performance (as shown in Table \ref{table:intro0} and \ref{table:intro}), we first investigate \emph{interleaved learning}, a method that intermingles the training data instead of simply concatenating, which ensures a model learns from all corpora evenly. We use this method as a baseline. Additionally, we explore two multi-domain learning methods: \emph{labeled learning} and \emph{multi-task labeled learning}. Labeled learning comes from a control technique in response generation \citep{li-persona,mt1,sigir-dual-learning}. Previous works focus on controlling persona and style, while our method controls corpus's information with the corpus embedding. Multi-task labeled learning is inspired by works of domain adaption \citep{multi-task-learning,polite,mt-survey}, where multiple losses from both the corpus classifier and response generator are minimized. To the best of our knowledge, this paper is the first that uses corpus embeddings on the open-domain response generation task for multiple corpora.

Furthermore, we propose a novel \emph{weighted learning} with Domain-specific Frequency (DF). DF is a word-level importance weight \citep{iw} that assigns different weights (importance) to the same words from different corpora. In the training process, we weight the loss of a model with DF, so that the model focuses on the most important words for a specific corpus.

For automatic evaluation metrics, we eliminate the stop words and use ROUGE-1 (precision, recall, F1) \citep{rouge} to measure the \textbf{relevance} of the generated responses. In addition, we adopt DF to see how relevant the generated response of a model is to a specific corpus. We will explain DF as an evaluation metric in Section \ref{df:explanation}. Results show that for overall performance, the best method (weighted learning) improves $27.4\%$ on precision, $45.5\%$ on recall, and $34.1\%$ on F1. Further, it has at least $20.0\%$ higher DF, stating that it uses more important words from the ``correct'' corpus. We also conduct an extensive human evaluation on 2400 generated responses. The human evaluation shows a highly significant ($p<0.001$) improvement on all of our proposed methods, especially the weighted learning method.

We summarize our work as follows: \begin{itemize}
    \item We explore the problem of training and evaluating on multiple corpora from different domains for open-domain response generation. The task is to make the conversational models generate relevant responses for {\bf each} corpus.
    \item We examine several multi-domain corpora learning methods for their ability to solve the proposed task.
    \item We propose Domain-specific Frequency (DF) as in weighted learning and as an evaluation metric. DF distinguishes important words for each corpus and helps a model to focus on these important words in the training process.
    
\end{itemize}

\section{Related Work}
\label{related_work}

\paragraph{Open-Domain Response Generation}

Recent work of open-domain response generation generally follows the work of \citet{ritter} where the task is treated as a machine translation task, and many of them use a Seq2Seq structure \citep{s2s} following previous work \citep{le,shang,sordoni}. In recent years, substantial improvements have been made \citep{iulian2,li-persona,transfertransfo}, and embeddings are used to control response generation on extra information such as persona \citep{li-persona}, profiles \citep{sigir-dual-learning}, coherence \citep{mix}, emotions \citep{emotion}, and dialogue attributes like response-relatedness \citep{stanford-human-evaluation}. However, there is a lack of work that uses embeddings to control response generation over multiple corpora. Our work follows the common models of open-domain conversational systems, while we study the problem of multiple corpora of different domains.

\paragraph{Multi-Domain Learning and Domain Adaption}

Multi-domain learning aims at making a conversational model learn from multiple domains to prevent the performance from degrading due to domain differences \citep{domain1}. There are two categories of solutions for multi-domain learning \citep{multi-domain}: (i) capturing domain-specific characteristics in the parameters \citep{specific}; (ii) capturing the relationship among different domains \citep{relation}. 

Some work of natural language generation and machine translation is related to multi-domain learning. \citet{multi-task-learning} and \citet{polite} use multi-task learning for domain adaption respectively on speaker-role and politeness. \citet{multi-domain-nlg} and \citet{pure-fine-tune} utilizes fine-tuning as a common way of domain adaption for language generator and style transferer. For machine translation, in order to deal with the mixed-domain parallel corpus, \citet{weighted-NMT} adjust the weights of target words in the training objective based on their relevance to different domains. We differ in that we propose DF and we deal with the response generation task. \citet{mix-fine-tune} propose mixed fine-tuning, which adds the out-of-domain pre-training data to the fine-tuning dataset, and they observe an improvement of performance. In this paper, we also mix small-scale fine-tuning datasets with out-of-domain training data, while the data we add is not necessarily used during pre-training. \citet{curriculum-learning} state that fine-tuning can be done by placing the corpus to be fine-tuned at the end of the entire corpus, which is an extension of curriculum learning proposed by \citet{curriculum}. We also explore how the order of multiple corpora influences the result, but our focus is on balancing performance. Recently, \citet{blend} investigated blending conversational skills with knowledge and empathy skills, where they mix 3 corpora. They focus on selecting appropriate skills and they propose a blended corpus with labels, while we focus on generating responses that are most relevant to a specific corpus.

\section{Base Models}
\label{model}

We use two base models: an LSTM Seq2Seq model with attention \citep{lstm,s2s,attention} and a pre-trained GPT-2 model \citep{gpt2}. The LSTM Seq2Seq model with attention is a common model for conversational systems \citep{li-persona,stanford-human-evaluation}, and the GPT2 model is a state-of-the-art model for the response generation task \citep{dialogpt,knowledge-grounded-pre-trained}.

The basic task of response generation is to predict the next word given the past and current words of the context and response, and to make the generated response as similar to the original response as possible. The task can be described as follows. Probability of response $Y$ given context $X$ is predicted as:
\begin{equation}\textstyle P(Y|X)=\prod_{t=1}^{n}P(y_t|y_1,\ldots,y_{t-1},X)\text{,}\end{equation}where $X=x_1,\ldots,x_m$ and $Y=y_1,\ldots,y_n$ is a context-response pair.

\subsection{LSTM Seq2Seq Model with Attention}\label{lstm}

We simplify an LSTM with attention unit as $\mathit{LSTM}^{*}$ since it is well introduced in previous work \citep{li-persona}. We calculate the hidden vector $h_t$ at step $t$ as:
\begin{equation}h_t = \mathit{LSTM}^{*}(h_{t-1},E(z_t))\text{,}\end{equation}where $h_{t-1}\in\R^{\mathit{dim}}$ is the hidden vector at step $t-1$, $\mathit{dim}$ is the dimension of hidden vectors, and $E(z_t)$ is the word embedding for word $z_t\in(x_1,\ldots,x_m,y_1,\ldots,y_{n-1})$. We apply dot multiple in the attention mechanism when calculating the context vector $c_t$: $$c_t=H\cdot(softmax(H^\top\cdot h_t))$$ where $H\in\R^{d\times m}$ is the concatenation of hidden vectors from the encoder. $c_t$ is input to the next step $t+1$ in the decoder. Each token's hidden vector $h_t$ in the decoder is combined with $c_t$ through a linear layer and an activation to predict the next token.

\subsection{GPT-2}

As for GPT-2, we follow the adaption of \citet{transfertransfo}. The transformer block of GPT-2 captures the relation of multiple words in one sentence, which largely follows \citet{transformer}. The hidden vector to be input to the transformer block is calculated as:\begin{equation}h_{0[t]}=E(X,Y_{[1:t]})+(E_0,E_1)+W_p\text{,}\end{equation}where $Y_{[1:t]}$ is $(y_1,\ldots,y_t)$, $E(X,Y_{[1:t]})$ is the sub-word embedding for context $X$ and response $Y_{[1:t]}$. $E_0$ and $E_1$ are dialogue-state embeddings, which tutor the model to distinguish between contexts and responses. $W_p$ is a pre-trained position embedding. The probability of the subword to generate is then calculated as:
\begin{align}
h_{[t]}&=\mathit{transformer\_block}(h_{0[t]})\\
P(y)_{t+1}&=\mathit{softmax}(E^\top(h_{[t]}))\text{,}\end{align}where $y\in V$, and $V$ stands for the sub-word vocabulary. We simplify the structure of transformer block as $\mathit{transformer\_block}$. In the block, a mask is filled in the attention matrix, which bans past words from attending to future words. This ensures that the model follows the traditional language modeling. The hidden vector of $t_{\text{th}}$ sub-word is used to generate the probability distribution for the vocabulary ($P(y),\ y\in V$) for $(t+1)_{\text{th}}$ sub-word. $E^\top$ means that the model uses the sub-word embeddings in calculating sub-word probabilities for generation \citep{mt2}.

\section{Proposed Methods}

\label{methodology}

\begin{figure*}[!ht]
\centering
\begin{subfigure}[c]{0.51\textwidth}
\centering
\includegraphics[width=\linewidth]{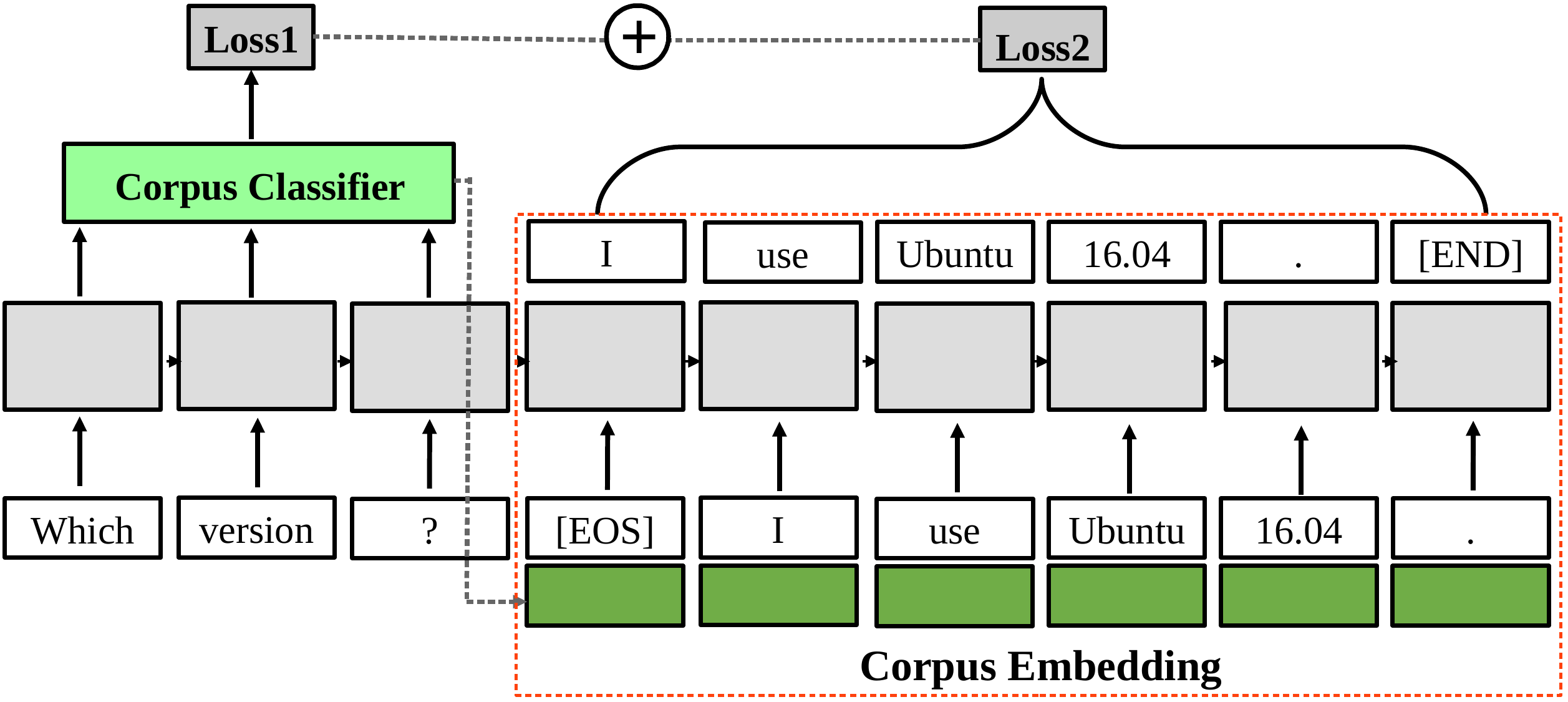}
\caption{Structure of multi-task labeled learning on LSTM model}\label{fig:mtl}
\includegraphics[width=0.95\linewidth]{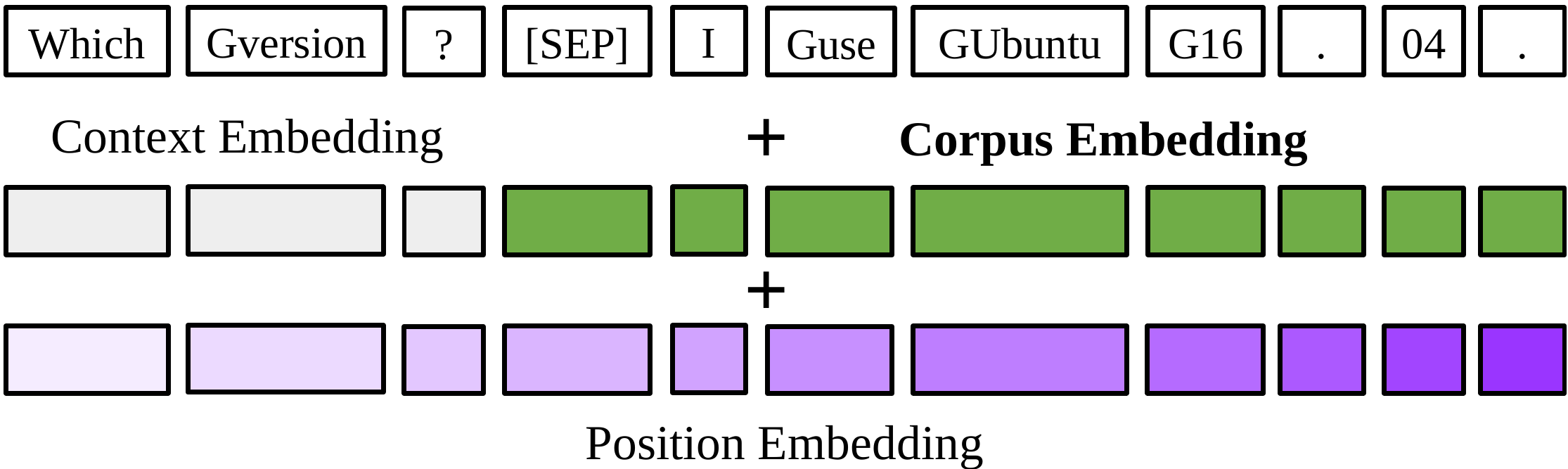}
\caption{Corpus embeddings with sub-word embeddings on GPT-2}\label{fig:gpt}
\end{subfigure}
\begin{subfigure}[c]{0.48\textwidth}
\includegraphics[width=\linewidth]{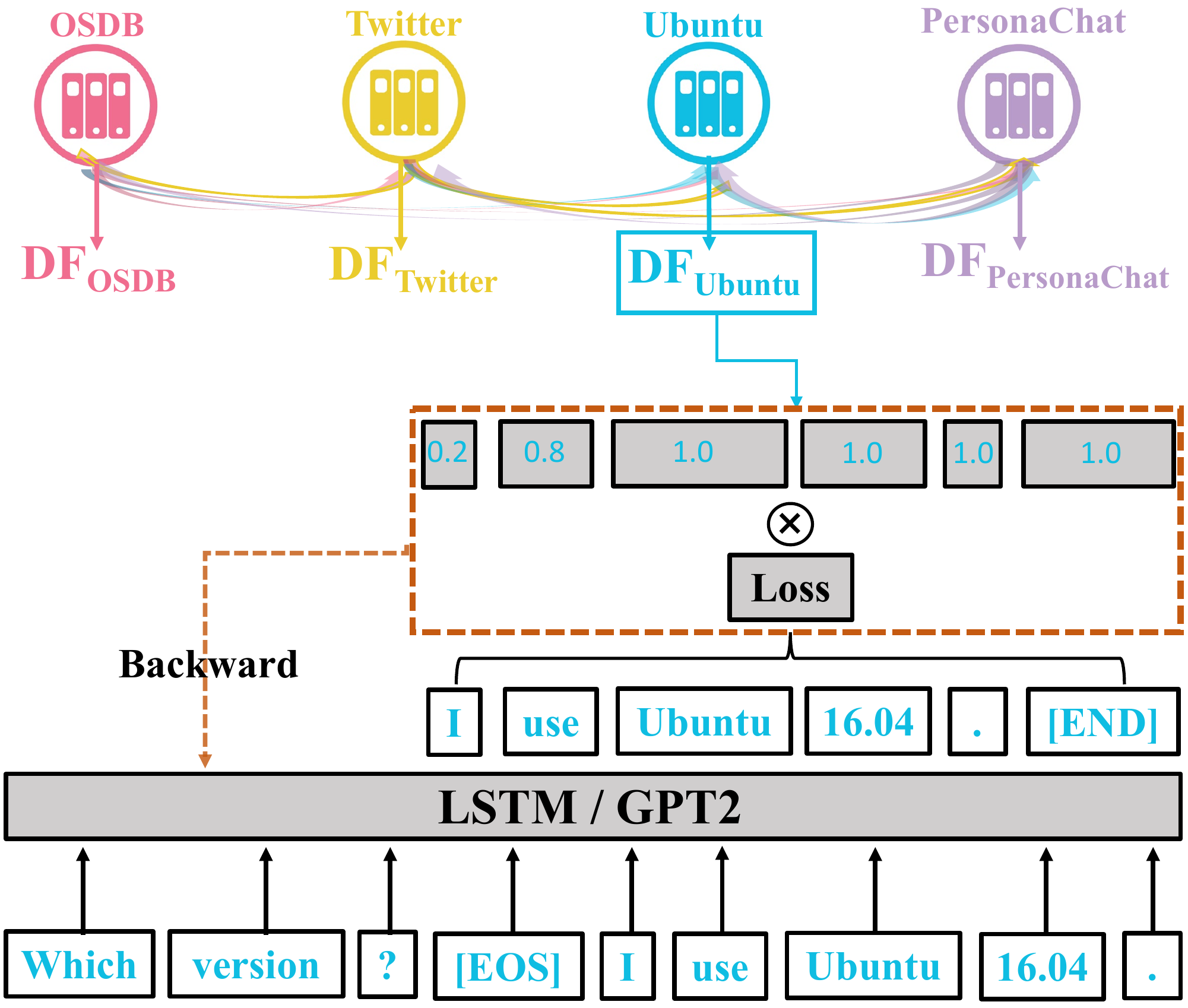}
\caption{Structure of weighted learning}\label{fig:weight}
\end{subfigure}
\caption{Adapted models with labeled learning, multi-task labeled learning and weighted learning}
\end{figure*}

\subsection{Interleaved Learning}\label{corpus}

Interleaving is a concept in cognitive psychology proven to be efficient for learning \citep{spaced-learning}: intermingling learning material of different topics helps students to gain better learning results than learning the material topic by topic. Previous work from machine learning also shows that training order greatly influences the performance \citep{curriculum}. When the training is conducted on a simple concatenation of multiple corpora, the model tends to concentrate on the last corpus \citep{curriculum-learning}. To address this issue, we propose interleaved learning as an alternative: each time we collect one context-response pair from each of the corpora, and we randomly shuffle them. For example, if there are 3 corpora $(a_1, a_2, ...), (b_1, b_2, ...), (c_1, c_2, ...)$ where $a_i, b_i$ and $c_i$ are context-response pairs, the resulting mixed corpus might be $(b_1, a_1, c_1, c_2, b_2, a_2, ...)$. Interleaved learning guarantees that the combined corpus is evenly distributed, which helps the model learn from multiple corpora evenly.

\subsection{Labeled Learning}\label{labeled_learning}

We propose our labeled learning as follows: each corpus is assigned a randomly initialized unique embedding, and the conversational model learns these embeddings together with conversations during the training period. We denote these embeddings as ``corpus embedding'', or $E_c$. A model captures each corpus's characteristics through the corpus embedding and uses it to control the generated responses. To know which corpus embedding to use, each context is labeled with which corpus it comes from, and these labels are provided to the model both in the training and generation period. We propose an approach for each of our base models for encoding corpus embeddings.

For the LSTM model, following \citet{li-persona}, we input the corpus embedding $E_c$ into the first layer of the decoder LSTM at every step, together with the response words. Calculation of a hidden vector $h_t$ in the decoder LSTM is then adapted to:\begin{equation}h_t = \mathit{LSTM}^{*}(h_{t-1},E(y_t),E_c)\text{.}\end{equation}The structure is illustrated in the dashed red rectangle of Figure \ref{fig:mtl}.

For the GPT-2 model, our method is based on \citet{transfertransfo}. Instead of two kinds of dialogue-state embeddings (context embedding $E_0$ and response embedding $E_1$), we replace the response embedding with corpus embeddings $E_c$. As a result, the model is aware of which corpus the response belongs. Calculation of a hidden vector to be input to the transformer block is adapted to:\begin{equation}h_{0[t]}=E(X,Y_{[1:t]})+(E_0,E_c)+W_p\text{.}\end{equation}The structure is illustrated in Figure \ref{fig:gpt}.

\subsection{Multi-Task Labeled Learning}

Labeled learning needs corpus labels for both training and generation processes. To avoid providing labels in the generation process, we combine multi-task learning with labeled learning on multiple corpora. Here, the conversational model has to predict by itself which corpus a context belongs to, which is expected to result in worse performance, but less information is required. In the encoder, we have a classifier layer that uses the sum of hidden vectors from the encoder ($\sum H$) to predict the corpus of a context. The loss of the classifier is calculated as:\begin{equation}\mathcal{L}_c =-\text{log}\left(\mathit{softmax}\left(\left(\sum H\right)\cdot W_{[c]}\right)\right)\text{,}\end{equation}where $W_{[c]}\in\R^{\mathit{dim}}$ is the part from the classifier layer for target corpus $c$. $\mathcal{L}_c$ is summed up with the loss from the response generator. The predicted corpus embedding is input into the decoder like labeled learning (see Section \ref{labeled_learning}). The simplified structure is illustrated in Figure \ref{fig:mtl}.

\subsection{Document-specific Frequency (DF)} We propose Domain-specific Frequency (DF) to measure how important a word is with respect to a different corpus under a collection of corpora. DF is used for weighted learning and evaluation. It is calculated as follows:\begin{align} 
f(w)_d &= \text{freq}(w)_d-{\min}_{v}\{\text{freq}(v)_d\}\label{eq:df1}\\
\text{df}(w)_d &= \begin{cases}
0 & f(w)_d=0\\
\frac{f(w)_d}{\sum_{d\in D} f(w)_d} & f(w)_d\neq0\label{eq:df2}\\
\end{cases}\\
\text{DF}(w)_d &= \frac{\text{df}(w)_d}{\max_{v}\{\text{df}(v)_d\}}\label{eq:df3}\text{,}\end{align}where $\text{freq}(w)_d$ is the relative frequency of a word $w$ in a corpus $d$, and $D$ represents the set of all corpora. It is easy to see from Equation \ref{eq:df2} that $\text{DF}(w)_d$ represents the importance of word $w$ for corpus $d$ compared to other corpora. For a word $w$ that frequently appears in corpus $d$ but seldom in other corpora (e.g.,  ``upgrade'' from Ubuntu corpus), $\sum_{d\in D} f(w)_d$ is close to $f(w)_d$, making $\text{DF}(w)_d$ approach $1$. A word that frequently appears in all corpora (e.g., ``I'', ``you'') is punished, resulting in a lower $\text{DF}(w)_d$. A word that seldom appears in corpus $d$ but frequently appears in other corpora (e.g., ``music'' seldom appears in Ubuntu corpus, but is common in other corpora) has the lowest $\text{DF}(w)_d$. Words that appear minimal times (e.g., once) in a corpus are ignored with Equation \ref{eq:df1}. Words that appear few times (e.g., twice or three times) are not dealt with, yet they are not of great influence in our experiments. We apply a normalization in the final step (Equation \ref{eq:df3}) to make DF$(w)_d$ of each corpus $d$ range from $0$ to $1$.

\label{df:explanation}

We show DF$(w)_{\text{Ubuntu}}$ and DF$(w)_{\text{PersonaChat}}$ of some words in Table \ref{table:df-edf}. We also show the results of TF-IDF (log normalization variant), a commonly used word importance weight, as a comparison. As expected, for the corpus Ubuntu and PersonaChat, most unique words $w$ have very different DF$(w)_{\text{Ubuntu}}$ and DF$(w)_{\text{PersonaChat}}$. Unique words of each corpus get the highest values for the corresponding corpus, like ``upgrade'' for the Ubuntu corpus and ``music'' for the PersonaChat corpus; these words receive the lowest values for \emph{incorrect} corpora, like ``upgrade'' for PersonaChat and ``music'' for Ubuntu. The stress on unique words makes DF more suitable for our task.

\begin{table}[!ht]
\centering
{\footnotesize
\centering
    \begin{tabular}{@{\hspace{0\tabcolsep}}c@{\hspace{0.5\tabcolsep}}| @{\hspace{0.5\tabcolsep}} c @{\hspace{0.5\tabcolsep}} c @{\hspace{0.5\tabcolsep}} | @{\hspace{0.5\tabcolsep}} c @{\hspace{0.5\tabcolsep}} c @{\hspace{0.5\tabcolsep}} | @{\hspace{0.5\tabcolsep}} c @{\hspace{0.5\tabcolsep}} c @{\hspace{0\tabcolsep}}}
    \toprule
        \multirow{2}{*}{Word} & \multicolumn{2}{c|@{\hspace{0.5\tabcolsep}}}{TF-IDF(\%)} & \multicolumn{2}{c|@{\hspace{0.5\tabcolsep}}}{DF(\%)} & \multicolumn{2}{c}{$\alpha$DF$_{(\alpha=100)}$}\\
        & \tiny{Ubuntu} & \tiny{PersonaChat} & \tiny{Ubuntu} & \tiny{PersonaChat} & \tiny{Ubuntu} & \tiny{PersonaChat} \\ 

    \hline
        i & 100.0 & 62.6 & 20.8 & 42.1 & 2.6 & 7.3\\
        to & 64.6 & 32.8 & 26.9 & 24.9 & 3.8 & 3.1\\
        it & 83.2 & 21.7 & 38.5 & 14.5 & 5.1 & 2.1\\
    \hline
        laptop & 5.4 & 0.2 & 89.8 & 4.5 & 76.0 & 1.0\\
        upgrade & 6.8 & 0.1 & 95.6 & 0.4 & 91.2 & 1.0\\
        file & 15.7 & 0.1 & 96.0 & 0.3 & 86.4 & 0\\
        windows & 12.2 & 0.1 & 97.1 & 0.1 & 86.3 & 1.0\\
        ubuntu & 27.5 & 0 & 99.9 & 0 & 99.5 & 0\\
        
    \hline
        teacher & 0.1 & 2.2 & 0.7 & 77.8 & 1.0 & 53.5\\
        music & 1.5 & 7.6 & 4.8 & 82.9 & 1.2 & 49.1\\
        travel & 0.1 & 3.1 & 0.3 & 88.9 & 1.0 & 57.1\\
        hobby & 0.1 & 1.6 & 0.6 & 94.3  & 1.1 & 81.7\\
        hiking & 0 & 1.5 & 0 & 97.6 & 0 & 91.8\\
    \bottomrule
    \end{tabular}}
    \caption{Normalized TF-IDF (\%), DF (\%) and $\alpha$DF of some words for Ubuntu and PersonaChat (more examples on other corpora can be found in Section \ref{app:df-example})}
    \label{table:df-edf}
\end{table}

\paragraph{Weighted Learning with DF} Weighted learning weights the loss of the predication $y'$ for each target word $w$ using $\text{DF}(w)_d$. In the training period, each context is labeled with the corpus $d$ it belongs to, so that the model can use the DF$(w)_d$ of the corresponding corpus. Here DF is calculated only on the training sets. In the generation step, corpus labels are not provided, so DF is not used. The loss is weighted as follows :\begin{equation}\mathcal{L}_{\text{weighted}}=\text{DF}(w)_d\cdot\left(-\text{log}\left(\mathit{softmax}(y'_w)\right)\right)\text{,}\end{equation}where $y'_w$ represents the model's predicted score for the target word $w$. With the weighted loss, the model concentrates on words that are important to the corpus of the current context, and focuses less on frequent words or words that are not important to the current corpus. The structure is illustrated in Figure \ref{fig:weight}.

\paragraph{Evaluation with DF} For the generated responses to be relevant to a specific corpus, they have to be similar to that corpus, which includes using important words of that corpus (e.g., responses generated for the Ubuntu corpus should have more technical words than other corpora). Thus, we propose DF as an evaluation metric that shows to what extent the generated responses use important words of the corresponding corpus. We want to decrease the influence of common words like ``i'', ``to'', etc., and thus address the important words. So we adopt exponential DF with $\alpha$ as the base ($\alpha$DF):
\begin{align}
\alpha\text{DF}(w)_d &= \begin{cases}
0 & \text{DF}(w)_d=0\\
\alpha^{\text{DF}(w)_d} & \text{DF}(w)_d\neq0\text{,}\\
\end{cases}
\end{align}where $\alpha$ is a constant. $\alpha\text{DF}(w)_d$ rescales $\text{DF}(w)_d$ by exponent with $\alpha$ as a base. In our experiments, we set $\alpha$ to be $100$, which transforms the range of the metric from $(0,1)$ to $(0,100)$. This makes the difference between high and low $\alpha$DF more significant than DF and gives a $100$-scale score. For each corpus $d\in D$, we average $\alpha\text{DF}(w)_d$ on word $w$ from the generated responses of each test set, which gives us $\alpha\text{DF}_d$ scores ($d\in D$) for each test set. Ideally, the generated responses of a specific corpus $d$ should have a higher $\alpha\text{DF}_d$ score and lower $\alpha\text{DF}_{\overline{d}}$ score ($\overline{d}\in\{d'\in D\mid d'\neq d\}$). For example, generated responses of the Ubuntu test set should have a higher $\alpha\text{DF}_{\text{Ubuntu}}$ score, while a lower $\alpha\text{DF}_{\overline{\text{Ubuntu}}}$ score ($\overline{\text{Ubuntu}}\in\{d'\in D\mid d'\neq\text{Ubuntu}\}$). $\alpha\text{DF}_d$ scores for responses from the original test sets are the standard scores.

We show $\alpha\text{DF}(w)_{\text{Ubuntu}}$ and $\alpha\text{DF}(w)_{\text{PersonaChat}}$ (calculated purely on test set) in Table \ref{table:df-edf}. As expected, $\alpha$DF has a more significant difference between important words and common words.

\paragraph{Is DF a Legal Evaluation Metric?} Although DF is used for both weighted learning and evaluation, we see DF as a suitable evaluation metric for our task and not biased in favor of weighted learning due to: 1) A word receives multiple DF values in the training process given the corpus that a context belongs to; 2) in the generation process, DF is never used. 3) In the evaluation process, DF can be calculated purely on the test sets. Note that since a word receives multiple DF values in the training step, it is equivalently likely for the model trained with weighted learning to be influenced by DF weights of \textbf{incorrect} corpus. Above all, in the evaluation step, if the trained model is influenced more by DF weights from the correct corpus, it already means that the model is good at distinguishing which corpus a given context is from, thus is suitable for our task.

\section{Experiment Setup}
\label{experiments}

\subsection{Datasets}

\paragraph{Data Collection} We collected 4 commonly used English corpora of different domains from the ParlAI platform \citep{parlai}: OpenSubtitles corpus (OSDB)\footnote{\url{http://www.opensubtitles.org/}} \citep{osdb}, Twitter corpus\footnote{\url{https://github.com/Marsan-Ma/chat_corpus/}} \citep{parlai}, Ubuntu chatlogs corpus \citep{ubuntu}\footnote{\url{https://github.com/rkadlec/ubuntu-ranking-dataset-creator}}
, and PersonaChat corpus \citep{persona-chat} from the NeurIPS 2018 ConvAI2 Challenge \citep{convai2}. Each corpus contains $250$K context-response pairs, as much as the size of the original PersonaChat used in ConvAI2 competition. This gives us $1$M context-response pairs in total. The corpus for training is a combination of these 4 corpora. For comparison, we have a single corpus--PersonaChat--trained on both base models. For testing, each of the 4 corpora has a test set of $30$K context-response pairs, which is the same size of the test set of PersonaChat.

The OpenSubtitles corpus (OSDB) is a noisy dataset of film subtitles. We removed films that belonged to genres that usually had few conversations, such as musical and documentary films. We regarded two neighboring sentences as a context-response pair following \citet{le}. The Twitter corpus contains one-turn dialogues extracted from Twitter. The original author has already cleaned it, so we only removed special symbols such as hashtags, Emojis, and @. The Ubuntu corpus contains dialogues about solving technical problems of Ubuntu. The PersonaChat corpus contains dialogues between two workers acting as specific personas; we focused on the dialogue part and ignored the persona part. This corpus allows us to compare our base models with state-of-the-art performance. These 4 corpora have very different characteristics, confirmed by the imbalanced performance of GPT-2 fine-tuned on a single corpus (see Table \ref{table:intro}).

\begin{table*}[!ht]
\centering
{\footnotesize
    \begin{tabular}{@{\hspace{0\tabcolsep}}l@{\hspace{0.5\tabcolsep}}|@{\hspace{0.7\tabcolsep}}l@{\hspace{0.5\tabcolsep}}|@{\hspace{0.5\tabcolsep}}c@{\hspace{0.6\tabcolsep}}c@{\hspace{0.8\tabcolsep}}c@{\hspace{0.6\tabcolsep}}|@{\hspace{0.5\tabcolsep}}c@{\hspace{0.6\tabcolsep}}c@{\hspace{0.7\tabcolsep}}c@{\hspace{0.6\tabcolsep}}|@{\hspace{0.5\tabcolsep}}c@{\hspace{0.6\tabcolsep}}c@{\hspace{\tabcolsep}}c@{\hspace{0.6\tabcolsep}}|@{\hspace{0.5\tabcolsep}}c@{\hspace{0.7\tabcolsep}}c@{\hspace{0.6\tabcolsep}}c@{\hspace{0.6\tabcolsep}}|@{\hspace{0.5\tabcolsep}}c@{\hspace{0.7\tabcolsep}}c@{\hspace{0.7\tabcolsep}}c@{\hspace{0\tabcolsep}}}
         & & \multicolumn{12}{c|@{\hspace{0.5\tabcolsep}}}{Test set} & \multicolumn{3}{c}{\quad}\\
        \multirow{2}{*}{Model} & \multirow{2}{*}{Corpus / Method} & \multicolumn{3}{c|@{\hspace{0.5\tabcolsep}}}{OSDB} & \multicolumn{3}{c|@{\hspace{0.5\tabcolsep}}}{Twitter} & \multicolumn{3}{c|@{\hspace{0.5\tabcolsep}}}{Ubuntu} & \multicolumn{3}{c|@{\hspace{0.5\tabcolsep}}}{PersonaChat} & \multicolumn{3}{c}{Overall}\\

        & & \tiny{Precision} & \tiny{Recall} & \tiny{F1} & \tiny{Precision} & \tiny{Recall} & \tiny{F1} & \tiny{Precision} & \tiny{Recall} & \tiny{F1} & \tiny{Precision} & \tiny{Recall} & \tiny{F1} & \tiny{Precision} & \tiny{Recall} & \tiny{F1} \\
    \hline\hline
        \multirow{6}{*}{LSTM} & PersonaChat (single) & 11.8&8.9&8.1&12.4&8.6&8.9&12.1&8.1&7.7&56.7&43.4&45.8&23.2&17.2&17.6\\
        & Concatenated & 11.0&7.7&7.2&15.7&10.9&11.4&36.5&17.8&20.1&57.7&\bf44.0&46.4&30.2&20.1&21.3\\
        \cline{2-17}
        & Interleaved & 24.1&10.1&11.7&24.3&12.5&14.9&58.4&24.9&29.6&56.1&41.5&44.3&40.7&22.3&25.1\\
        & Labeled & 23.9&10.1&11.3&24.5&\bf13.2&15.5&61.6&26.5&31.6&56.4&43.0&45.4&41.6&23.2&26.0\\
        & Multi-task Labeled & 23.2&9.6&11.1&23.2&12.3&14.5&56.4&23.8&28.3&53.2&40.6&42.7&39.0&21.6&24.2\\
        & Weighted & \bf26.6&\bf11.9&\bf13.4&\bf29.7&12.2&\bf15.6&\bf78.4&\bf35.2&\bf41.2&\bf62.4&42.5&\bf47.1&\bf49.3&\bf25.5&\bf29.3\\
    \hline\hline
        \multirow{6}{*}{GPT-2} & PersonaChat (single) & 15.0&12.4&10.8&19.6&13.2&13.9&24.8&16.2&15.5&70.0&57.1&58.8&32.4&24.7&24.7\\
        & Concatenated & 17.4&14.1&12.6&24.5&16.4&17.2&35.0&22.5&22.4&66.8&55.4&56.3&35.9&27.1&27.1\\
        \cline{2-17}
        & Interleaved & 40.0&20.5&22.3&31.0&17.9&20.1&81.7&38.1&44.3&68.7&56.2&57.6&55.3&33.2&36.1\\
        & Labeled & 38.6&19.9&21.6&31.4&\bf19.4&21.1&84.2&38.4&45.0&\bf70.7&\bf57.2&\bf59.0&56.2&33.7&36.7\\
        & Multi-task Labeled & 38.4&19.8&21.4&31.2&18.6&20.6&80.9&37.8&43.8&68.0&56.0&57.3&54.6&33.0&35.8\\
        & Weighted & \bf41.9&\bf21.2&\bf23.4&\bf39.9&18.4&\bf22.3&\bf86.8&\bf43.3&\bf48.6&69.0&53.2&55.8&\bf59.4&\bf34.0&\bf37.5\\
    \end{tabular}}
    \caption{Precision, recall and F1 of ROUGE-1 (\textperthousand) for baselines and proposed methods fine-tuned on 4 corpora (stop words eliminated)}
    \label{table:score}
\end{table*}

\begin{table*}[!ht]
\centering
{\footnotesize
    \begin{tabular}{@{\hspace{0\tabcolsep}}l@{\hspace{0.5\tabcolsep}}|l@{\hspace{0.8\tabcolsep}}|c@{\hspace{0.8\tabcolsep}}c@{\hspace{\tabcolsep}}|c@{\hspace{0.5\tabcolsep}}c@{\hspace{0.8\tabcolsep}}|c@{\hspace{0.5\tabcolsep}}c@{\hspace{0.8\tabcolsep}}|c@{\hspace{0.5\tabcolsep}}c@{\hspace{0.8\tabcolsep}}|c@{\hspace{0.5\tabcolsep}}c@{\hspace{0.8\tabcolsep}}|c@{\hspace{0.5\tabcolsep}}c@{\hspace{0.8\tabcolsep}}|c@{\hspace{0.5\tabcolsep}}c}
         & & \multicolumn{14}{c}{Test set} \\
        \multirow{4}{*}{Model} & \multirow{4}{*}{Corpus / Method} & \multicolumn{4}{c|}{OSDB} & \multicolumn{4}{c|}{Twitter} & \multicolumn{4}{c|}{Ubuntu} & \multicolumn{2}{c}{PersonaChat} \\

        & & \multicolumn{2}{c|}{\scriptsize{OSDB}} & \multicolumn{2}{c|}{\color{gray}\scriptsize{PersonaChat}} & \multicolumn{2}{c|}{\scriptsize{Twitter}} & \multicolumn{2}{c|}{\color{gray}\scriptsize{PersonaChat}} & \multicolumn{2}{c|}{\scriptsize{Ubuntu}} & \multicolumn{2}{c|}{\color{gray}\scriptsize{PersonaChat}} & \multicolumn{2}{c}{\scriptsize{PersonaChat}} \\
        
         & & \multicolumn{14}{c}{\tiny{$\alpha$DF Calculated From: }} \\
        
        & & \scriptsize{Train} & \scriptsize{Test} & \color{gray}\scriptsize{Train} & \color{gray}\scriptsize{Test} & \scriptsize{Train} & \scriptsize{Test} & \color{gray}\scriptsize{Train} & \color{gray}\scriptsize{Test} & \scriptsize{Train} & \scriptsize{Test} & \color{gray}\scriptsize{Train} & \color{gray}\scriptsize{Test} & \scriptsize{Train} & \scriptsize{Test}  \\

    \hline\hline
        \multicolumn{2}{c|}{Test Set (Standard Score)} & \bf7.0&\bf9.7&\color{gray}3.6&\color{gray}3.7&
        \bf9.1&\bf11.0&\color{gray}3.6&\color{gray}3.8&
        \bf19.4&\bf23.2&\color{gray}2.7&\color{gray}2.8&
        \bf9.5&\bf12.0\\
    \hline\hline
        \multirow{6}{*}{LSTM} & PersonaChat (single)& 2.9&3.4&\color{gray}9.2&\color{gray}9.9&
        2.8&3.4&\color{gray}8.6&\color{gray}9.2&
        2.7&3.1&\color{gray}8.6&\color{gray}9.1&
        11.9&12.6\\
        & Concatenated & 2.9&3.3&\color{gray}7.6&\color{gray}8.6&
        3.6&4.3&\color{gray}8.0&\color{gray}8.7&
        7.6&7.7&\color{gray}5.6&\color{gray}6.0&
        12.5&13.6\\
    \cline{2-16}
        & Interleaved & 3.9&4.1&\color{gray}5.0&\color{gray}5.3&
        4.7&4.9&\color{gray}4.1&\color{gray}4.5&
        11.8&11.3&\color{gray}3.7&\color{gray}4.0&
        11.5&12.5\\
        & Labeled & 3.9&4.2&\color{gray}5.0&\color{gray}5.3& 5.0&5.3&\color{gray}3.9&\color{gray}4.3&
        12.5&11.8&\color{gray}3.4&\color{gray}3.8&
        12.1&13.1\\
        & Multi-task Labeled & 3.8&4.0&\color{gray}5.0&\color{gray}5.4&
        4.5&4.7&\color{gray}4.1&\color{gray}4.5&
        11.2&10.7&\color{gray}3.8&\color{gray}4.1&
        11.4&12.6\\
        & Weighted & \bf5.6&\bf6.3&\color{gray}4.1&\color{gray}4.5&
        \bf9.9&\bf10.1&\color{gray}3.8&\color{gray}4.3&
        \bf27.7&\bf25.4&\color{gray}2.7&\color{gray}3.0&
        \bf17.7&\bf18.3\\
    \hline\hline
        \multirow{6}{*}{GPT-2} & PersonaChat (single) & 2.8&3.2&\color{gray}10.5&\color{gray}11.1&
        2.9&3.3&\color{gray}9.5&\color{gray}9.8&
        4.1&4.6&\color{gray}8.3&\color{gray}8.4&
        12.9&13.7\\
        & Concatenated & 3.1&3.6&\color{gray}8.8&\color{gray}9.4&
        3.3&3.9&\color{gray}8.2&\color{gray}8.7&
        6.5&7.1&\color{gray}7.0&\color{gray}7.4&
        12.1&13.0\\
    \cline{2-16}
        & Interleaved & 4.9&5.8&\color{gray}4.8&\color{gray}5.0&
        4.6&5.1&\color{gray}4.4&\color{gray}4.7&
        15.7&16.0&\color{gray}3.1&\color{gray}3.4&
        12.1&12.9\\
        & Labeled & 4.9&5.8&\color{gray}4.8&\color{gray}5.0&
        4.7&5.2&\color{gray}4.1&\color{gray}4.3&
        16.7&17.0&\color{gray}2.9&\color{gray}3.2&
        12.4&13.1\\
        & Multi-task Labeled & 4.8&5.7&\color{gray}4.8&\color{gray}5.1&
        4.6&5.1&\color{gray}4.4&\color{gray}4.6&
        15.5&15.8&\color{gray}3.1&\color{gray}3.4&
        12.1&12.9\\
        & Weighted & \bf6.0&\bf7.5&\color{gray}4.1&\color{gray}4.4&
        \bf8.1&\bf8.8&\color{gray}3.7&\color{gray}4.1&
        \bf25.7&\bf24.4&\color{gray}2.4&\color{gray}2.6&
        \bf16.0&\bf17.1\\

    \end{tabular}}
    \caption{$\alpha$DF$_d$ scores for generated responses from multiple corpora. The columns ``train'' indicate train-set-$\alpha$DF$_d$. The columns ``test'' indicate test-set-$\alpha$DF$_d$.}
    \label{table:df}
\end{table*}

\subsection{Training and Decoding}

We used Pytorch \citep{pytorch} to implement the LSTM Seq2Seq model with attention and the pre-trained GPT-2 models. For GPT-2, we adapted our model from the implementation of the HuggingFace team\footnote{\url{https://huggingface.co/}.}. The LSTM model has 4 layers and the dimension is 512. The training procedure was with a batch size of 256, learning rate of 1.0, dropout rate of 0.2, and gradient clip threshold of 5. The vocabulary size is 50000. GPT-2 has 12 layers, 12 heads, and the dimension is 768, the same as the pre-trained model. The training procedure was with Adam and we adopted a similar setup as \citet{transfertransfo}: the batch size was 32, learning rate was $6\times10^{-5}$, $\beta_1=0.9$, $\beta_2=0.999$, L2 weight decay set to $0.01$, learning rate linearly decreased to zero at the end. We followed these hyper-parameters to ensure state-of-the-art performance for the base models. We use the same hyper-parameters for both base models and models with our proposed methods, so the proposed methods work slightly (but not much) worse than it should be. This is to avoid the extra improvement caused by hyper-parameters. We pre-trained the LSTM model on 3 large-scale corpora (OSDB, Twitter and Ubuntu) with interleaved learning until converging. GPT-2 is already pre-trained, so we directly used it for fine-tuning (details about pre-training convergence can be found in Section \ref{app:converge}). For decoding, we adopted greedy decoding for all the models to ensure an equal condition.

\subsection{Evaluation}\label{df-bias}

For automatic metrics, to measure the \textbf{relevance} of the generated responses, we eliminated punctuation and stop words, and adopted Rouge-1\footnote{We used implementation from \url{https://github.com/google-research/google-research/tree/master/rouge}.} (precision, recall, F1) as multi-grams become meaningless without stop words. However, Rouge-1 compares the generated responses with the golden ones, while there is never a standard response for any context, so in addition to Rouge, we use $\alpha$DF score that shows to what extent the generated responses use important words of the corresponding corpus, as stated in Section \ref{df:explanation}. Due to the limitation of automatic evaluation methods \citep{evaluation-review}, we also conduct an extensive human evaluation on the relevance of generated responses to contexts (see Section \ref{sec:human} for details).

\begin{table*}[ht!]
\centering
\small
    \begin{tabular}{l| c c c c |c}
    \toprule
    \multicolumn{1}{c|}{Model $\backslash$ Corpus} & OSDB & Twitter & Ubuntu & PersonaChat & Overall \\
    \hline
    PersonaChat (single) & 1.53 & 1.43 & 1.21 & 2.09 & 1.56\\
    Concatenated & 1.67 & 1.71 & 1.60 & 2.16 & 1.78\\
    \hline
    Interleaved & 2.04 & 1.89 & 2.18 & 2.24 & 2.09\\
    Labeled & 2.10 & 2.10 & 2.32 & 2.24 & 2.19\\
    Multi-task Labeled & 2.05 & 1.98 & 2.11 & 2.24 & 2.10\\
    Weighted & \bf 2.40 & \bf 2.45 & \bf 2.61 & \bf 2.47 & \bf 2.48\\
    \bottomrule
    \end{tabular}
\caption{Average scores of human evaluation for GPT-2 based models on each corpus}\label{human}
\end{table*}

\begin{table*}[!ht]
\centering
{\footnotesize
    \begin{tabular}{@{\hspace{0.5\tabcolsep}}l@{\hspace{0.8\tabcolsep}}|@{\hspace{0.6\tabcolsep}}l@{\hspace{\tabcolsep}}l|@{\hspace{0.8\tabcolsep}}l@{\hspace{\tabcolsep}}l@{\hspace{0\tabcolsep}}l@{\hspace{\tabcolsep}}l@{\hspace{0.5\tabcolsep}}}
    \toprule
    Model $\backslash$ Model & \ PersonaChat & \ Concatenated & \ \ Interleaved & \ \ \ Labeled & Multi-Task Labeled & Weighted \\
    \hline
    PersonaChat & \quad\ \ \color{gray}$1.00$ & \qquad $\backslash$ & \qquad\ $\backslash$ & \qquad\ $\backslash$ & \qquad\quad $\backslash$ & \quad\ \ $\backslash$\\
    
    Concatenated & $2.54\times10^{-7**}$ & \quad\ \color{gray}$1.00$ & \qquad\ $\backslash$ & \qquad\ $\backslash$ & \qquad\quad $\backslash$ & \quad\ \ $\backslash$\\
    
    \hline
    
    Interleaved & $4.71\times10^{-34**}$ & $2.09\times10^{-12**}$ & \quad\ \color{gray}$1.00$ & \qquad\ $\backslash$ & \qquad\quad$\backslash$ & \quad\ \ $\backslash$\\
    
    Labeled & $1.08\times10^{-46**}$ & $9.41\times10^{-21**}$ & $1.18\times10^{-2*}$ & \quad\ \ \color{gray}$1.00$ & \qquad\quad$\backslash$ & \quad\ \ $\backslash$\\
    
    Multi-task Labeled & $6.65\times10^{-35**}$ & $6.96\times10^{-13**}$ & $8.86\times10^{-1}$ & $1.17\times10$ & \qquad\ \color{gray}$1.00$ & \quad\ \ $\backslash$\\
    
    Weighted & $1.65\times10^{-103**}$ & $2.86\times10^{-63**}$ & $6.54\times10^{-26**}$ & $1.59\times10^{-15**}$ & \ \ \ $2.01\times10^{-25**}$ & \quad\color{gray}$1.00$\\
    \bottomrule
    \end{tabular}}
\caption{P-value for t-test on overall human evaluation scores of GPT-2 based models, $^{**} \ p<0.001$}\label{significance}
\end{table*}

\section{Results}
\label{results}

Our base models achieve perplexity scores of $28.9$ (LSTM model) and $19.6$ (GPT-2) on the test set of the PersonaChat dataset from the ConvAI2 competition when fine-tuned with the single PersonaChat corpus (more details can be found in Section \ref{app:score_origin}). These results would likely advance the models to the second round in the competition.

Table \ref{table:score} shows that models trained with our proposed methods gain better performance on Rouge than baselines. Baselines concentrate on the last trained corpus (PersonaChat), while with the proposed methods, performance is more balanced on multiple corpora. Weighted learning has the best overall performance on all metrics, and it performs especially well on the Ubuntu corpus, indicating that it might be good at distinguishing the unique technical words from the Ubuntu corpus. Labeled learning is the second best with stable improvement from interleaved learning, indicating that the corpus embeddings function as expected. Multi-task labeled learning has slightly worse performance than interleaved learning, indicating that predicting the corpus of a contexts is not easy, and wrong predictions result in worse performance.

Table \ref{table:df} shows $\alpha$DF$_d$ scores for generated responses of each corpus. Full results can be found in Section \ref{app:df-full}. We use both $\alpha$DF$_d$ calculated purely on the train set (train-set-$\alpha$DF) and $\alpha$DF$_d$ calculated purely on the test set (test-set-$\alpha$DF). The black scores are scores for the corresponding corpus (we expect high scores for these parts), while the grey scores are scores for non-related corpus--PersonaChat (we expect low scores for these parts). Note that scores for different corpora are in different scales. From the table, we can see that train-set-DF scores and test-set-DF scores are similar, and weighted learning always has the highest score, indicating that weighted learning distinguishes well which corpus a context comes from. Labeled learning is the second best, indicating that the learned corpus embeddings help the model to use more important words of the corresponding corpus. Compared to the concatenated corpus, the improvement is at least $20\%$, while the decrease in PersonaChat is just $9\%$ at most.

\subsection{Human Evaluation}\label{sec:human}

We conducted a human evaluation on all GPT-2 models: base models and models adapted with our proposed methods. We randomly picked $2400$ responses: $400$ different contexts evenly from 4 corpora with $6$ responses generated by each of our models. 3 judges\footnote{Similar to previous work like \citet{dialogpt}, we have 3 judges. We have one random worker from \url{https://www.mturk.com/worker}, one bachelor student, and one graduate student. An example of the mTurk interface can be found in Section \ref{app:human_evaluation_system}.} are asked to pick the most and the least relevant response(s) for the given context. The most relevant response(s) are given score $3$, the least relevant response(s) are given score $1$, and the other(s) are given score $2$. Table \ref{human} shows the overall scores of all GPT-2 based models. Table \ref{significance} shows the p-value for the t-test conducted between every two models. The overall scores of our proposed methods are all highly significantly ($p<0.001$) higher than the concatenated models, especially the weighted learning method.

\subsection{Response Examples}\label{responses}

The generated responses from better methods are more relevant to the corresponding corpus, while worse methods cannot distinguish contexts from different corpora (e.g., they may answer any questions in a ``PersonaChat'' way). To show an intuition of the difference among our proposed methods, we present some response examples generated by GPT-2 in Section \ref{app:response-example}.  

\subsection{Possible Limitations}

Our proposed methods are meant to be able to work in most models, which is why we choose the most common conversational models as our base models. However, there are many variants of conversational models focusing on different aspects, such as integrating knowledge, avoiding dull responses, keeping the speech style, etc. We cannot ensure that our methods work for all of these variant models. Also, dialogues are always multi-turn, while we focus on a simpler task: single-turn response generation. Furthermore, the methods are trained and evaluated on English corpora. There can be a limitation on applying the methods to other languages.

\section{Conclusions}
\label{conclusion}

We have experimented with 4 methods--interleaved learning (baseline), labeled learning, multi-task labeled learning, and weighted learning--to help common open-domain conversational systems generate relevant responses for multiple corpora of different domains. We adopted Rouge (precision, recall, F1) for auto evaluation. In addition, we used DF to evaluate how well a model uses relevant words for a corresponding corpus. We also did an extensive human evaluation. Our results show significant improvement in performance for our proposed methods, especially weighted learning. Future work of multi-turn response generation is potential. We have focused on one-turn response generation, while dialogue is naturally multi-turn so further research is needed.


\section*{Acknowledgements}

This paper is funded by the collaborative project of DNB ASA and Norwegian University of Science and Technology (NTNU). We also received assist on computing resources from the IDUN cluster of NTNU \citep{IDUN}. We would like to thank Aria Rahmati, Zhirong Yang (Norwegian Research Council, 287284) and {\"O}zlem {\"O}zg{\"o}bek for their helpful comments.

\bibliography{references}
\bibliographystyle{acl_natbib}
\newpage

\appendix
\onecolumn


\section{Comparison among TF-IDF, DF and $\alpha$DF for $4$ corpora on more example words}\label{app:df-example}

\bigskip

\begin{table}[!ht]
\centering
{\small
\centering
    \begin{tabular}{@{\hspace{0\tabcolsep}}c@{\hspace{0.5\tabcolsep}}|c@{\hspace{\tabcolsep}}c@{\hspace{\tabcolsep}}c@{\hspace{0.5\tabcolsep}}c@{\hspace{0.5\tabcolsep}}|c@{\hspace{1.5\tabcolsep}}c@{\hspace{1.5\tabcolsep}}c@{\hspace{0.5\tabcolsep}}c@{\hspace{0.5\tabcolsep}}|c@{\hspace{1.2\tabcolsep}}c@{\hspace{1.2\tabcolsep}}c@{\hspace{0.5\tabcolsep}}c}
    \toprule
        \multirow{2}{*}{Word} & \multicolumn{4}{c|}{TF-IDF(\%)} & \multicolumn{4}{c|}{DF(\%)} & \multicolumn{4}{c}{$\alpha$DF$_{(\alpha=100)}$}\\
        & \scriptsize{OSDB} & \scriptsize{Twitter} & \scriptsize{Ubuntu} & \scriptsize{PersonaChat}
        & \scriptsize{OSDB} & \scriptsize{Twitter} & \scriptsize{Ubuntu} & \scriptsize{PersonaChat}
        & \scriptsize{OSDB} & \scriptsize{Twitter} & \scriptsize{Ubuntu} & \scriptsize{PersonaChat}\\ 

    \hline
        i & 91.39 & 100.00 & 100.00 & 62.63 & 21.40 & 15.68 & 20.80 & 42.12 & 2.62 & 2.01 & 2.59 & 7.32\\
        to & 54.46 & 77.55 & 64.59 & 32.80 & 24.85 & 23.40 & 26.87 & 24.89 & 3.00 & 2.88 & 3.76 & 3.08\\
        it & 61.77 & 74.10 & 83.20 & 21.74 & 25.02 & 22.02 & 38.49 & 14.46 & 3.44 & 2.67 & 5.11 & 2.13\\
    \hline
        sword & \cellcolor{r}0.64 & 0.17 & 0.01 & 0.08 & \cellcolor{r}68.37 & 13.74 & 0.26 & 17.63 & \cellcolor{r}63.29 & 1.37 & 1.00 & 1.15\\
        forgive & \cellcolor{r}2.41 & 0.48 & 0.16 & 0.06 & \cellcolor{r}75.35 & 14.37 & 5.44 & 4.84 & \cellcolor{r}50.96 & 1.58 & 1.19 & 1.05\\
        hurry & \cellcolor{r}5.21 & 0.52 & 0.09 & 0.08 & \cellcolor{r}88.39 & 6.67 & 1.48 & 3.45 & \cellcolor{r}63.53 & 1.32 & 1.15 & 1.04\\
        darling & \cellcolor{r}2.54 & 0.39 & 0.00 & 0.01 & \cellcolor{r}90.88 & 8.42 & 0.11 & 0.58 & \cellcolor{r}57.10 & 1.45 & 0 & 1.21\\
        explain & \cellcolor{r}1.27 & 0.00 & 0.00 & 0.11 & \cellcolor{r}91.33 & 0 & 0 & 8.67 & \cellcolor{r}94.14 & 0 & 0 & 1.06\\
        
    \hline
        tax & 0.21 & \cellcolor{y}2.52 & 0.05 & 0.09 & 6.77 & \cellcolor{y}87.06 & 1.09 & 5.07 & 1.28 & \cellcolor{y}71.26 & 1.05 & 1.04\\
        liberal & 0.03 & \cellcolor{y}1.71 & 0.01 & 0.10 & 2.06 & \cellcolor{y}88.19 & 0.25 & 9.50 & 1.21 & \cellcolor{y}59.65 & 0 & 1.38\\
        vote & 0.41 & \cellcolor{y}6.08 & 0.10 & 0.11 & 6.07 & \cellcolor{y}90.68 & 0.78 & 2.47 & 1.12 & \cellcolor{y}80.22 & 1.02 & 1.09\\
        trump & 0.04 & \cellcolor{y}18.66 & 0.00 & 0.13 & 0.11 & \cellcolor{y}99.16 & 0.00 & 0.73 & 1.00 & \cellcolor{y}96.63 & 0 & 1.03\\
        hillary & 0.05 & \cellcolor{y}8.61 & 0.00 & 0.01 & 0.42 & \cellcolor{y}99.53 & 0 & 0.05 & 0 & \cellcolor{y}99.38 & 0 & 1.01\\
        
    \hline
        laptop & 0.10 & 0.40 & \cellcolor{b}5.39 & 0.15 & 1.33 & 4.37 & \cellcolor{b}89.88 & 4.42 & 1.07 & 1.22 & \cellcolor{b}76.02 & 1.01\\
        upgrade & 0.03 & 0.47 & \cellcolor{b}6.85 & 0.03 & 0.24 & 3.75 & \cellcolor{b}95.63 & 0.37 & 1.01 & 1.06 & \cellcolor{b}91.24 & 1.03\\
        file & 0.64 & 0.55 & \cellcolor{b}15.65 & 0.05 & 2.29 & 1.44 & \cellcolor{b}96.02 & 0.26 & 1.11 & 1.04 & \cellcolor{b}86.36 & 0\\
        windows & 0.33 & 0.44 & \cellcolor{b}12.18 & 0.06 & 1.09 & 1.37 & \cellcolor{b}97.13 & 0.41 & 1.04 & 1.10 & \cellcolor{b}86.33 & 1.01\\
        ubuntu & 0.00 & 0.01 & \cellcolor{b}27.47 & 0.00 & 0 & 0.01 & \cellcolor{b}99.99 & 0 & 0 & 1.01 & \cellcolor{b}99.48 & 0\\
    \hline
        music & 1.90 & 3.29 & 1.53 & \cellcolor{p}7.66 & 4.01 & 8.20 & 4.84 & \cellcolor{p}82.94 & 1.18 & 1.40 & 1.23 & \cellcolor{p}49.14\\
        teacher & 1.48 & 0.74 & 0.07 & \cellcolor{p}2.20 & 14.53 & 7.01 & 0.68 & \cellcolor{p}77.78 & 1.39 & 1.32 & 1.01 & \cellcolor{p}53.49\\
        travel & 0.42 & 0.91 & 0.05 & \cellcolor{p}3.07 & 3.91 & 6.89 & 0.28 & \cellcolor{p}88.92 & 1.27 & 1.36 & 1.01 & \cellcolor{p}57.15\\
        hobby & 0.10 & 0.27 & 0.04 & \cellcolor{p}1.56 & 1.94 & 3.03 & 0.57 & \cellcolor{p}94.46 & 1.13 & 1.00 & 1.09 & \cellcolor{p}81.71\\
        hiking & 0.03 & 0.09 & 0.00 & \cellcolor{p}1.52 & 0.85 & 1.45 & 0 & \cellcolor{p}97.70 & 0 & 1.09 & 0 & \cellcolor{p}91.76\\
    \bottomrule
    \end{tabular}}
    \caption{Normalized TF-IDF (\%), DF (\%) and $\alpha$DF of more example words for 4 corpora}
\end{table}

Example words are divided into five blocks. The first block has frequent words in all corpora, the second block has unique words from OSDB, the third block has unique words from Twitter, the fourth block has unique words from Ubuntu, and the fifth block has unique words from PersonaChat. The values of the corresponding corpus are marked with different colors.

From this table, it is clear that the commonly used word importance weight, TF-IDF, is not suitable for our task. This is due to the vast range of frequency, which leads to a relatively small penalty for IDF (Inversed Document Frequency) over words with too large TF (Term Frequency).

\bigskip

\section{Convergence time of pre-training LSTM model on large-scale corpora}\label{app:converge}

\begin{figure*}[!h]
\centering
\begin{subfigure}[b]{0.45\textwidth}
\includegraphics[width=\linewidth]{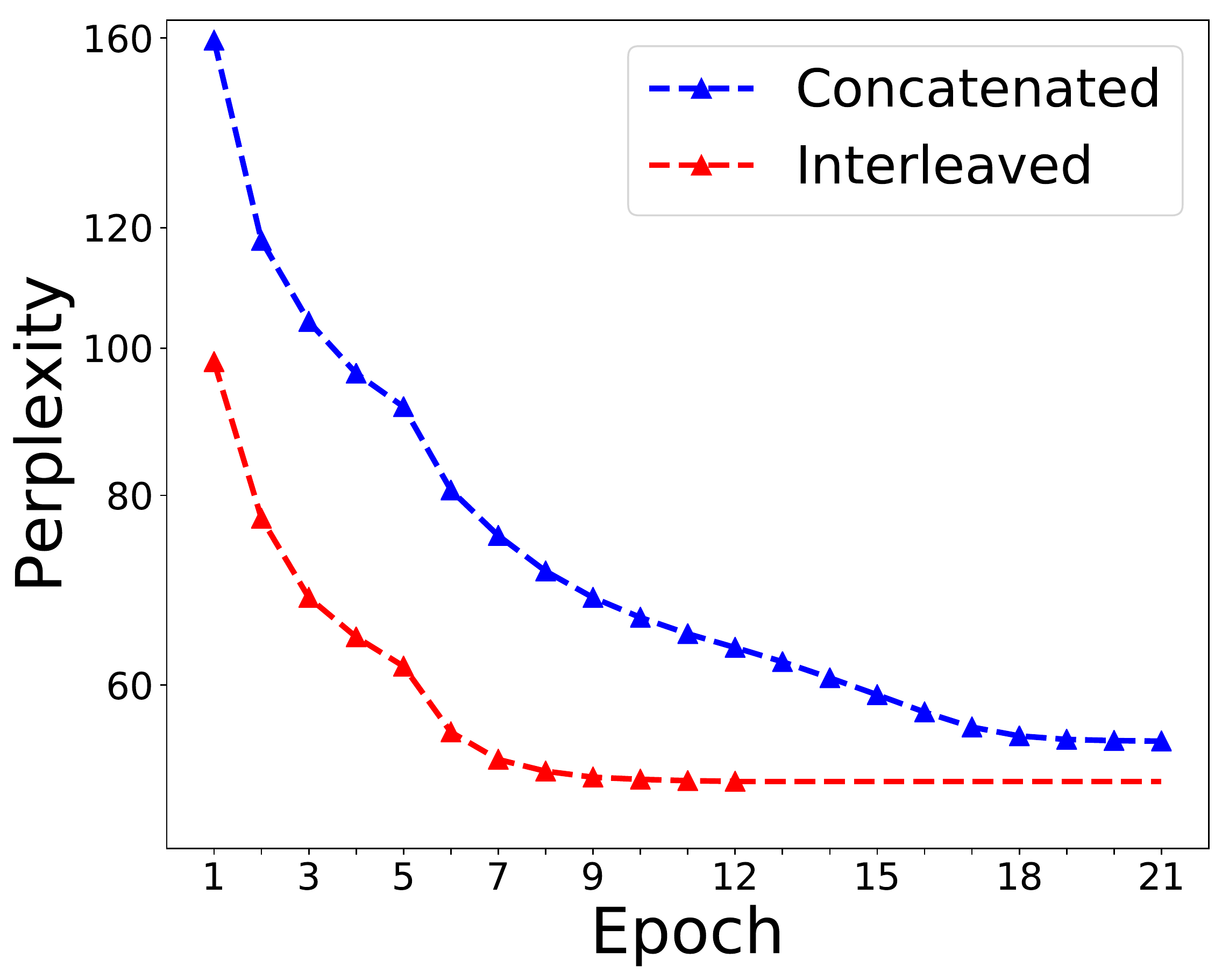}
\caption{Overall perplexity of 3 corpora (OpenSubtitles, Twitter, Ubuntu) per epoch}
\end{subfigure}
\hspace*{0.1in}
\begin{subfigure}[b]{0.45\textwidth}
\includegraphics[width=\linewidth]{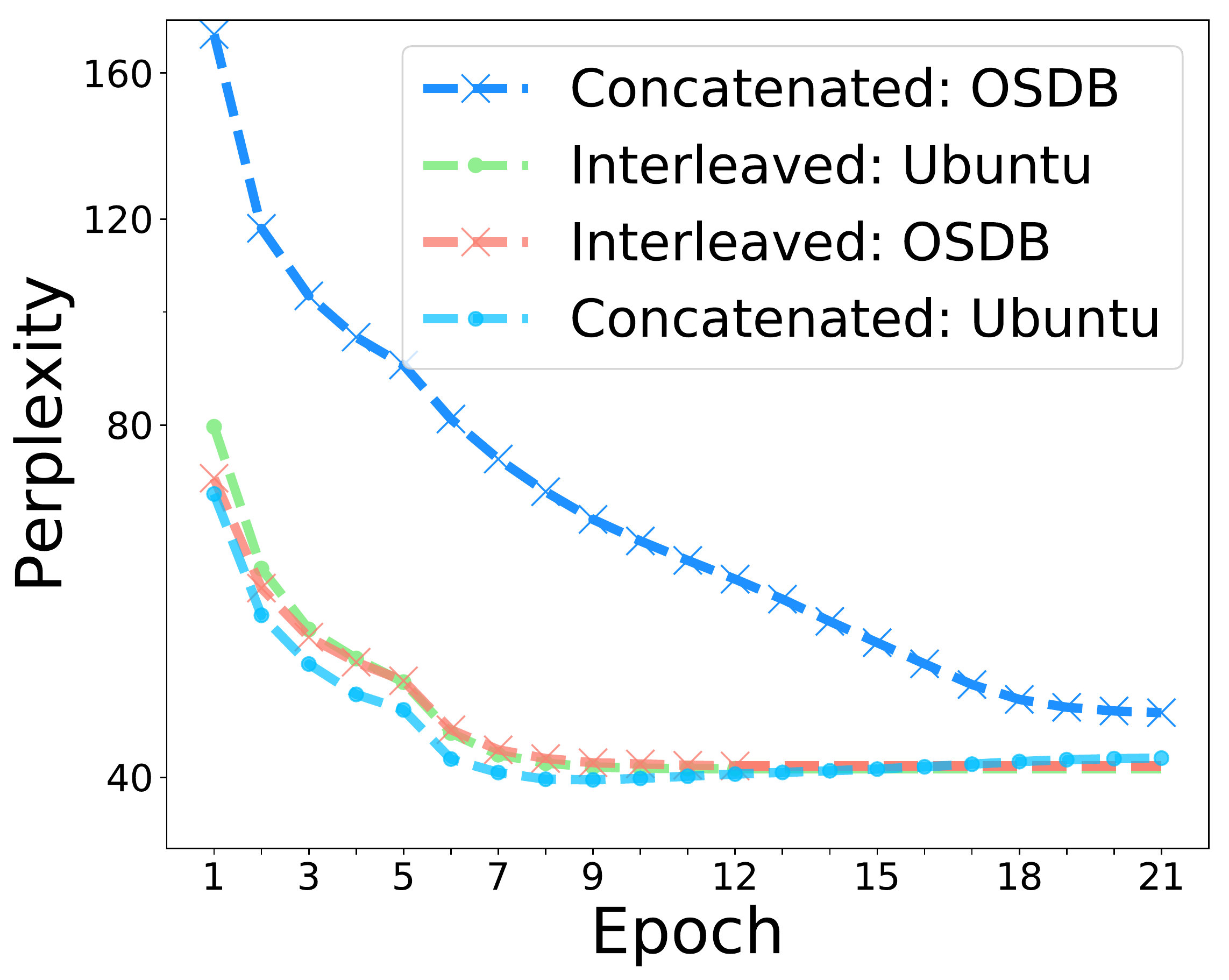}
\caption{Perplexity of OSDB corpus and Ubuntu corpus per epoch}
\end{subfigure}

\caption{Convergence time of pre-training LSTM on large-scale corpora}
\end{figure*}

In the pre-training period, it takes 21 epochs for the concatenated corpus to converge on the base LSTM model, while only 12 epochs with interleaved learning, which is $43\%$ shorter. When trained on the concatenated corpus in the order of OSDB $\rightarrow$ Twitter $\rightarrow$ Ubuntu, it takes 20 epochs for the perplexity on OSDB and Ubuntu to be balanced, while with interleaved learning, it takes less than one epoch. For concatenated corpus, the performance of the Ubuntu corpus is sacrificed in order to balance the performance of the two corpora, which results in worse overall performance.


\section{Results of automatic evaluation with stop words}\label{app:score_origin}

\begin{table*}[!ht]
\centering
{\footnotesize
    \begin{tabular}{@{\hspace{0\tabcolsep}}l@{\hspace{0.5\tabcolsep}}|l@{\hspace{0.8\tabcolsep}}|@{\hspace{0.9\tabcolsep}}c@{\hspace{0.6\tabcolsep}}c@{\hspace{0.8\tabcolsep}}c@{\hspace{0.9\tabcolsep}}|@{\hspace{0.9\tabcolsep}}c@{\hspace{0.6\tabcolsep}}c@{\hspace{0.7\tabcolsep}}c@{\hspace{0.9\tabcolsep}}|@{\hspace{0.9\tabcolsep}}c@{\hspace{0.6\tabcolsep}}c@{\hspace{\tabcolsep}}c@{\hspace{0.9\tabcolsep}}|@{\hspace{0.9\tabcolsep}}c@{\hspace{0.7\tabcolsep}}c@{\hspace{0.6\tabcolsep}}c@{\hspace{0.9\tabcolsep}}|@{\hspace{0.9\tabcolsep}}c@{\hspace{0.7\tabcolsep}}c@{\hspace{0.7\tabcolsep}}c@{\hspace{0\tabcolsep}}}
         & & \multicolumn{12}{c|@{\hspace{0.9\tabcolsep}}}{Test set} & \multicolumn{3}{c}{\quad}\\
        \multirow{2}{*}{Model} & \multirow{2}{*}{Corpus / Method} & \multicolumn{3}{c|@{\hspace{0.9\tabcolsep}}}{OSDB} & \multicolumn{3}{c|@{\hspace{0.9\tabcolsep}}}{Twitter} & \multicolumn{3}{c|@{\hspace{0.9\tabcolsep}}}{Ubuntu} & \multicolumn{3}{c|@{\hspace{0.9\tabcolsep}}}{PersonaChat} & \multicolumn{3}{c}{Overall}\\

        & & \scriptsize{Perp} & \scriptsize{BLEU} & \scriptsize{F1} & \scriptsize{Perp} & \scriptsize{BLEU} & \scriptsize{F1} & \scriptsize{Perp} & \scriptsize{BLEU} & \scriptsize{F1} & \scriptsize{Perp} & \scriptsize{BLEU} & \scriptsize{F1} & \scriptsize{Perp} & \scriptsize{BLEU} & \scriptsize{F1} \\
    \hline\hline
        \multirow{6}{*}{LSTM} & PersonaChat (single) & 109.8&4.8&6.5 & 191.9&5.4&6.3 & 116.9&4.8&6.8 & 28.9&13.1&15.0 & 47.0&7.0&8.7\\
        & Concatenated & 57.0&\bf4.8&6.3 & 111.4&5.9&6.1 & 50.0&5.1&6.8 & 27.8&13.2&15.1 & 36.8&\bf7.2&8.6\\
        & Interleaved & 41.3&3.7&\bf6.7 & 89.3&6.0&7.6 & 43.1&5.1&8.7 & 27.9&12.8&15.0 & 34.3&6.9&9.5\\
        \cline{2-17}
        & Labeled & \bf40.5&3.2&6.6 & \bf87.0&\bf6.2&7.6 & \bf42.6&\bf5.3&\bf8.8 & \bf27.1&\bf13.2&\bf15.2 & \bf33.4&7.0&\bf9.6\\
        & Multi-task Labeled & 41.7&3.5&6.6 & 89.7&6.1&\bf7.7 & 43.5&5.0&8.6 & 27.8&12.6&14.8 & 34.3&6.8&9.4\\
        & Weighted & 46.1&3.6&6.6 & 102.5&4.6&6.7 & 49.4&3.8&6.6 & 32.8&11.4&15.0 & 39.9&5.8&8.7\\
    \hline\hline
        \multirow{6}{*}{GPT-2} & PersonaChat (single) & 478.8&4.9&6.7 & 159.6&5.5&6.7 & 264.7&5.1&7.7 & 19.6&14.1&16.2 & 44.7&7.3&9.3\\
        & Concatenated & 392.8&\bf5.0&6.9 & 110.7&5.8&7.0 & 199.2&5.8&8.5 & 19.0&13.9&16.0 & 40.1&\bf7.6&9.6\\
        & Interleaved & 26.6&4.3&7.4 & 54.8&5.8&7.4 & 28.1&5.7&9.2 & 19.2&14.0&16.1 & 23.7&7.4&10.0\\
        \cline{2-17}
        & Labeled & \bf26.5&4.2&7.3 & \bf54.1&\bf5.9&\bf7.6 & \bf27.7&5.7&9.2 & \bf18.9&\bf14.1&\bf16.3 & \bf23.5&7.5&\bf10.1\\
        & Multi-task Labeled & 26.9&4.1&7.2 & 55.4&5.8&7.5 & 38.5&\bf5.8&\bf9.4 & 20.7&14.0&16.1 & 25.1&7.4&10.1\\
        & Weighted & 29.6&4.3&\bf7.5 & 64.1&5.1&7.4 & 44.1&4.1&7.0 & 23.4&13.0&15.7 & 28.4&6.6&9.4\\
    \end{tabular}}
    \caption{Perplexity, BLEU (\%) and F1 (\%) scores for baselines and proposed methods fine-tuned on 4 corpora (\textbf{with} stop words). BLEU is from NLTK sentence BLEU}
\end{table*}

Models of labeled, multi-task labeled and weighted learning do not have the best hyper-parameters, but the same hyper-parameters as the base models. Their perplexity is slightly worse than it should be.

The results of the single corpus PersonaChat trained with the LSTM model confirm our concern on a small fine-tuning corpus. The LSTM model is pre-trained on OSDB, Twitter and Ubuntu; however, the performance for the 3 corpora greatly decreases after fine-tuning.

The automatic evaluation with stop words is not good for measuring relevance, since stop words are taken too much into account. See BLEU and F1 scores of PersonChat (single) and weighted learning as an example. Models trained on PersonaChat (single) cannot answer Ubuntu technical questions \textbf{at all}, yet they receive better scores than weighted learning. But once the stop words are removed, the scores of weighted learning surplus PersonaChat (single) a lot.


\section{Additional Results of automatic evaluation without stop words}\label{app:result2}

\begin{table*}[!ht]
\centering
{\footnotesize
    \begin{tabular}{@{\hspace{0\tabcolsep}}l@{\hspace{0.5\tabcolsep}}|@{\hspace{0.5\tabcolsep}}l@{\hspace{0.8\tabcolsep}}|@{\hspace{0.6\tabcolsep}}c@{\hspace{0.6\tabcolsep}}c@{\hspace{0.8\tabcolsep}}c@{\hspace{0.9\tabcolsep}}|@{\hspace{0.6\tabcolsep}}c@{\hspace{0.6\tabcolsep}}c@{\hspace{0.7\tabcolsep}}c@{\hspace{0.9\tabcolsep}}|@{\hspace{0.6\tabcolsep}}c@{\hspace{0.6\tabcolsep}}c@{\hspace{\tabcolsep}}c@{\hspace{0.9\tabcolsep}}|@{\hspace{0.6\tabcolsep}}c@{\hspace{0.7\tabcolsep}}c@{\hspace{0.6\tabcolsep}}c@{\hspace{0.9\tabcolsep}}|@{\hspace{0.6\tabcolsep}}c@{\hspace{0.7\tabcolsep}}c@{\hspace{0.7\tabcolsep}}c@{\hspace{0\tabcolsep}}}
         & & \multicolumn{12}{c|@{\hspace{0.6\tabcolsep}}}{Test set} & \multicolumn{3}{c}{\quad}\\
        \multirow{2}{*}{Model} & \multirow{2}{*}{Corpus / Method} & \multicolumn{3}{c|@{\hspace{0.6\tabcolsep}}}{OSDB} & \multicolumn{3}{c|@{\hspace{0.6\tabcolsep}}}{Twitter} & \multicolumn{3}{c|@{\hspace{0.6\tabcolsep}}}{Ubuntu} & \multicolumn{3}{c|@{\hspace{0.6\tabcolsep}}}{PersonaChat} & \multicolumn{3}{c}{Overall}\\

        & & \tiny{BLEU} & \tiny{ROUGE} & \tiny{DF-F1} & \tiny{BLEU} & \tiny{ROUGE} & \tiny{DF-F1} & \tiny{BLEU} & \tiny{ROUGE} & \tiny{F1} & \tiny{BEU} & \tiny{ROUGE} & \tiny{DF-F1} & \tiny{BLEU} & \tiny{ROUGE} & \tiny{DF-F1} \\
    \hline\hline
        \multirow{6}{*}{LSTM} & PersonaChat (single) & 5.2&8.1&6.2&5.7&8.9&5.0&4.5&7.7&4.8&34.2&45.8&44.6&12.4&17.6&15.2\\
        & Concatenated & 4.5&7.2&5.6&7.4&11.4&8.8&11.6&20.1&17.4&\bf34.6&46.4&44.2&14.5&21.3&19.0\\
        & Interleaved & 6.5&11.7&9.9&8.6&14.9&12.6&17.1&29.6&28.4&32.4&44.3&43.2&16.1&25.1&23.5\\
        \cline{2-17}
        & Labeled & 6.2&11.3&9.7&\bf9.1&15.5&12.6&18.1&31.6&30.7&33.5&45.4&43.8&16.7&26.0&24.2\\
        & Multi-task Labeled & 6.2&11.1&9.5&8.4&14.5&11.7&16.0&28.3&27.2&31.5&42.7&41.9&15.5&24.2&22.6\\
        & Weighted & \bf7.6&\bf13.4&\bf12.2&7.6&\bf15.6&\bf18.7&\bf24.2&\bf41.2&\bf44.1&33.2&\bf47.1&\bf46.9&\bf18.2&\bf29.3&\bf30.5\\
    \hline\hline
        \multirow{6}{*}{GPT-2} & PersonaChat (single) & 7.1&10.8&9.2&8.7&13.9&10.5&8.8&15.5&12.2&45.0&58.8&56.8&17.4&24.7&22.2\\
        & Concatenated & 8.4&12.6&11.0&10.8&17.2&13.7&13.4&22.4&23.3&43.0&56.3&55.7&18.9&27.1&25.9\\
        & Interleaved & 14.0&22.3&21.3&12.2&20.1&19.3&25.8&44.3&48.3&44.2&57.6&58.0&24.0&36.1&36.7\\
        \cline{2-17}
        & Labeled & 13.6&21.6&20.5&\bf13.1&21.1&20.3&25.8&45.0&49.6&\bf45.1&\bf59.0&\bf59.6&24.4&36.7&37.5\\
        & Multi-task Labeled & 13.4&21.4&20.4&12.7&20.6&20.1&25.4&43.8&47.6&44.0&57.3&57.4&23.9&35.8&36.4\\
        & Weighted & \bf14.5&\bf23.4&\bf23.4&11.9&\bf22.3&\bf25.2&\bf29.2&\bf48.6&\bf52.5&42.4&55.8&57.6&\bf24.5&\bf37.5&\bf39.7\\
    \end{tabular}}
    \caption{BLEU (\textperthousand), ROUGE (\textperthousand) and DF-F1 (\textperthousand) scores for baselines and proposed methods fine-tuned on 4 corpora (\textbf{without} stop words). DF-F1 is ROUGE F1 weighted by test-set $\alpha$DF}
\end{table*}

\newpage


\section{Full results of $\alpha$DF for generated responses from multiple corpora}\label{app:df-full}

\bigskip

\begin{table}[!ht]
\centering
\begin{subtable}[!ht]{\textwidth}
\centering
{\small
    \begin{tabular}{l|l|cc|cc|cc|cc}
        \multirow{4}{*}{Model} & \multirow{4}{*}{Corpus / Method} & \multicolumn{8}{c}{Test set: OSDB} \\

        & & \multicolumn{2}{c|}{\footnotesize{OSDB}} & \multicolumn{2}{c|}{\color{gray}\footnotesize{Twitter}} & \multicolumn{2}{c|}{\color{gray}\footnotesize{Ubuntu}} & \multicolumn{2}{c|}{\color{gray}\footnotesize{PersonaChat}}\\
        
         & & \multicolumn{8}{c}{\tiny{$\alpha$DF Calculated From: }} \\
        
        & & \scriptsize{Train} & \scriptsize{Test} & \color{gray}\scriptsize{Train} & \color{gray}\scriptsize{Test} & \color{gray}\scriptsize{Train} & \color{gray}\scriptsize{Test} & \color{gray}\scriptsize{Train} & \color{gray}\scriptsize{Test} \\

    \hline\hline
        \multicolumn{2}{c|}{Test Set (Standard Score)} & \bf7.01&\bf9.66&\color{gray}3.75&\color{gray}3.75&\color{gray}2.82&\color{gray}2.86&\color{gray}3.59&\color{gray}3.75\\
    \hline\hline
        \multirow{6}{*}{LSTM} & PersonaChat (single)& 2.92&3.40&\color{gray}2.40&\color{gray}2.82&\color{gray}2.27&\color{gray}2.51&\color{gray}9.18&\color{gray}9.91\\
        & Concatenated & 2.92&3.35&\color{gray}2.49&\color{gray}2.94&\color{gray}2.41&\color{gray}2.71&\color{gray}7.65&\color{gray}8.55\\
        & Interleaved & 3.88&4.13&\color{gray}2.45&\color{gray}2.54&\color{gray}2.89&\color{gray}2.87&\color{gray}4.98&\color{gray}5.31\\
    \cline{2-10}
        & Labeled & 3.94&4.16&\color{gray}2.37&\color{gray}2.44&\color{gray}2.71&\color{gray}2.70&\color{gray}5.01&\color{gray}5.34\\
        & Multi-task Labeled & 3.78&4.02&\color{gray}2.41&\color{gray}2.49&\color{gray}2.91&\color{gray}2.88&\color{gray}5.02&\color{gray}5.36\\
        & Weighted & \bf5.60&\bf6.29&\color{gray}2.65&\color{gray}2.84&\color{gray}2.89&\color{gray}2.84&\color{gray}4.14&\color{gray}4.47\\
    \hline\hline
        \multirow{6}{*}{GPT-2} & PersonaChat (single) & 2.76&3.15&\color{gray}2.30&\color{gray}2.66&\color{gray}2.24&\color{gray}2.51&\color{gray}10.53&\color{gray}11.09\\
        & Concatenated & 3.07&3.59&\color{gray}2.52&\color{gray}2.96&\color{gray}2.30&\color{gray}2.55&\color{gray}8.75&\color{gray}9.35\\
        & Interleaved & 4.86&5.78&\color{gray}2.63&\color{gray}2.67&\color{gray}2.69&\color{gray}2.66&\color{gray}4.77&\color{gray}5.04\\
        \cline{2-10}
        & Labeled & 4.86&5.77&\color{gray}2.61&\color{gray}2.66&\color{gray}2.67&\color{gray}2.64&\color{gray}4.76&\color{gray}5.04\\
        & Multi-task Labeled & 4.81&5.70&\color{gray}2.60&\color{gray}2.64&\color{gray}2.69&\color{gray}2.65&\color{gray}4.83&\color{gray}5.1\\
        & Weighted & \bf6.02&\bf7.46&\color{gray}2.71&\color{gray}2.83&\color{gray}2.47&\color{gray}2.48&\color{gray}4.12&\color{gray}4.38\\

    \end{tabular}}
    \caption{$\alpha$DF$_d$ scores for generated responses from OSDB}
    \label{table:df-osdb}
\end{subtable}

\bigskip

\bigskip

\begin{subtable}[!ht]{\textwidth}
\centering
{\small
    \begin{tabular}{l|l|cc|cc|cc|cc}
        \multirow{4}{*}{Model} & \multirow{4}{*}{Corpus / Method} & \multicolumn{8}{c}{Test set: Twitter} \\

        & & \multicolumn{2}{c|}{\color{gray}\footnotesize{OSDB}} & \multicolumn{2}{c|}{\footnotesize{Twitter}} & \multicolumn{2}{c|}{\color{gray}\footnotesize{Ubuntu}} & \multicolumn{2}{c|}{\color{gray}\footnotesize{PersonaChat}}\\
        
         & & \multicolumn{8}{c}{\tiny{$\alpha$DF Calculated From: }} \\
        
        & & \color{gray}\scriptsize{Train} & \color{gray}\scriptsize{Test} & \scriptsize{Train} & \scriptsize{Test} & \color{gray}\scriptsize{Train} & \color{gray}\scriptsize{Test} & \color{gray}\scriptsize{Train} & \color{gray}\scriptsize{Test} \\

    \hline\hline
        \multicolumn{2}{c|}{Test Set (Standard Score)} & \color{gray}3.97&\color{gray}4.07&\bf9.07&\bf11.01&\color{gray}3.24&\color{gray}3.40&\color{gray}3.64&\color{gray}3.80\\
    \hline\hline
        \multirow{6}{*}{LSTM} & PersonaChat (single)&
        \color{gray}2.79&\color{gray}3.21&2.78&3.36&\color{gray}2.35&\color{gray}2.59&\color{gray}8.60&\color{gray}9.18\\
        & Concatenated & \color{gray}2.62&\color{gray}3.12&3.55&4.31&\color{gray}2.30&\color{gray}2.71&\color{gray}7.97&\color{gray}8.69\\
        & Interleaved & \color{gray}3.28&\color{gray}3.68&4.66&4.95&\color{gray}3.11&\color{gray}3.34&\color{gray}4.11&\color{gray}4.51\\
    \cline{2-10}
        & Labeled & \color{gray}3.30&\color{gray}3.68&4.97&5.27&\color{gray}3.00&\color{gray}3.24&\color{gray}3.89&\color{gray}4.26\\
        & Multi-task Labeled & \color{gray}3.31&\color{gray}3.68&4.47&4.73&\color{gray}3.14&\color{gray}3.36&\color{gray}4.08&\color{gray}4.49\\
        & Weighted & \color{gray}3.10&\color{gray}3.62&\bf9.92&\bf10.10&\color{gray}2.79&\color{gray}3.01&\color{gray}3.79&\color{gray}4.30\\
    \hline\hline
        \multirow{6}{*}{GPT-2} & PersonaChat (single)&
        \color{gray}2.74&\color{gray}3.04&2.87&3.33&\color{gray}2.45&\color{gray}2.66&\color{gray}9.47&\color{gray}9.77\\
        & Concatenated & \color{gray}2.87&\color{gray}3.28&3.32&3.94&\color{gray}2.41&\color{gray}2.65&\color{gray}8.21&\color{gray}8.68\\
        & Interleaved & \color{gray}3.42&\color{gray}3.67&4.59&5.08&\color{gray}3.05&\color{gray}3.13&\color{gray}4.39&\color{gray}4.68\\
        \cline{2-10}
        & Labeled & \color{gray}3.48&\color{gray}3.74&4.66&5.16&\color{gray}3.08&\color{gray}3.19&\color{gray}4.06&\color{gray}4.35\\
        & Multi-task Labeled & \color{gray}3.41&\color{gray}3.66&4.63&5.11&\color{gray}3.08&\color{gray}3.15&\color{gray}4.37&\color{gray}4.65\\
        & Weighted & \color{gray}3.58&\color{gray}4.01&\bf8.13&\bf8.84&\color{gray}2.59&\color{gray}2.79&\color{gray}3.68&\color{gray}4.07\\

    \end{tabular}}
    \caption{$\alpha$DF$_d$ scores for generated responses from Twitter}
    \label{table:df-twitter}
\end{subtable}

\bigskip

\bigskip

\begin{subtable}[!ht]{\textwidth}
\centering
{\small
    \begin{tabular}{l|l|cc|cc|cc|cc}
        \multirow{4}{*}{Model} & \multirow{4}{*}{Corpus / Method} & \multicolumn{8}{c}{Test set: Ubuntu} \\

        & & \multicolumn{2}{c|}{\color{gray}\footnotesize{OSDB}} & \multicolumn{2}{c|}{\color{gray}\footnotesize{Twitter}} & \multicolumn{2}{c|}{\footnotesize{Ubuntu}} & \multicolumn{2}{c|}{\color{gray}\footnotesize{PersonaChat}}\\
        
         & & \multicolumn{8}{c}{\tiny{$\alpha$DF Calculated From: }} \\
        
        & & \color{gray}\scriptsize{Train} & \color{gray}\scriptsize{Test} &  \color{gray}\scriptsize{Train} & \color{gray}\scriptsize{Test} & \scriptsize{Train} & \scriptsize{Test} & \color{gray}\scriptsize{Train} & \color{gray}\scriptsize{Test} \\

    \hline\hline
        \multicolumn{2}{c|}{Test Set (Standard Score)} & \color{gray}2.69&\color{gray}2.74&\color{gray}2.96&\color{gray}2.85&\bf19.36&\bf23.20&\color{gray}2.67&\color{gray}2.78\\
    \hline\hline
        \multirow{6}{*}{LSTM} & PersonaChat (single)&
        \color{gray}2.71&\color{gray}3.28&\color{gray}2.41&\color{gray}2.89&2.74&3.06&\color{gray}8.55&\color{gray}9.09\\
        & Concatenated & \color{gray}2.61&\color{gray}2.89&\color{gray}2.27&\color{gray}2.53&7.60&7.74&\color{gray}5.59&\color{gray}5.99\\
        & Interleaved & \color{gray}2.91&\color{gray}3.19&\color{gray}2.30&\color{gray}2.36&11.78&11.27&\color{gray}3.70&\color{gray}4.01\\
    \cline{2-10}
        & Labeled & \color{gray}3.03&\color{gray}3.38&\color{gray}2.28&\color{gray}2.36&12.46&11.75&\color{gray}3.45&\color{gray}3.75\\
        & Multi-task Labeled & \color{gray}2.91&\color{gray}3.17&\color{gray}2.30&\color{gray}2.35&11.19&10.72&\color{gray}3.77&\color{gray}4.09\\
        & Weighted & \color{gray}2.16&\color{gray}2.84&\color{gray}2.05&\color{gray}2.16&\bf27.73&\bf25.42&\color{gray}2.68&\color{gray}3.01\\
    \hline\hline
        \multirow{6}{*}{GPT-2} & PersonaChat (single)&
        \color{gray}2.60&\color{gray}2.85&\color{gray}2.31&\color{gray}2.64&4.12&4.64&\color{gray}8.27&\color{gray}8.42\\
        & Concatenated & \color{gray}2.67&\color{gray}3.03&\color{gray}2.45&\color{gray}2.82&6.54&7.10&\color{gray}7.04&\color{gray}7.37\\
        & Interleaved & \color{gray}2.73&\color{gray}3.05&\color{gray}2.22&\color{gray}2.37&15.67&16.02&\color{gray}3.08&\color{gray}3.41\\
        \cline{2-10}
        & Labeled & \color{gray}2.68&\color{gray}3.03&\color{gray}2.17&\color{gray}2.35&16.73&17.02&\color{gray}2.90&\color{gray}3.24\\
        & Multi-task Labeled & \color{gray}2.73&\color{gray}3.06&\color{gray}2.22&\color{gray}2.37&15.45&15.78&\color{gray}3.12&\color{gray}3.44\\
        & Weighted & \color{gray}2.26&\color{gray}2.56&\color{gray}2.16&\color{gray}2.28&\bf25.73&\bf24.42&\color{gray}2.37&\color{gray}2.60\\

    \end{tabular}}
    \caption{$\alpha$DF$_d$ scores for generated responses from Ubuntu}
    \label{table:df-ubuntu}
\end{subtable}
\end{table}

\newpage

\begin{table}[!ht]
\ContinuedFloat
\centering
\begin{subtable}[!ht]{\textwidth}
\centering
{\small
    \begin{tabular}{l|l|cc|cc|cc|cc}
        \multirow{4}{*}{Model} & \multirow{4}{*}{Corpus / Method} & \multicolumn{8}{c}{Test set: PersonaChat} \\

        & & \multicolumn{2}{c|}{\color{gray}\footnotesize{OSDB}} & \multicolumn{2}{c|}{\color{gray}\footnotesize{Twitter}} & \multicolumn{2}{c|}{\color{gray}\footnotesize{Ubuntu}} & \multicolumn{2}{c|}{\footnotesize{PersonaChat}}\\
        
         & & \multicolumn{8}{c}{\tiny{$\alpha$DF Calculated From: }} \\
        
        & & \color{gray}\scriptsize{Train} & \color{gray}\scriptsize{Test} &  \color{gray}\scriptsize{Train} & \color{gray}\scriptsize{Test} & \color{gray}\scriptsize{Train} & \color{gray}\scriptsize{Test} & \scriptsize{Train} & \scriptsize{Test} \\

    \hline\hline
        \multicolumn{2}{c|}{Test Set (Standard Score)} & \color{gray}3.32&\color{gray}3.23&\color{gray}3.18&\color{gray}3.04&\color{gray}2.67&\color{gray}2.69&\bf9.45&\bf12.00\\
    \hline\hline
        \multirow{6}{*}{LSTM} & PersonaChat (single)&
        \color{gray}2.59&\color{gray}3.02&\color{gray}2.31&\color{gray}2.73&\color{gray}2.15&\color{gray}2.35&11.86&12.62\\
        & Concatenated & \color{gray}2.47&\color{gray}2.84&\color{gray}2.29&\color{gray}2.76&\color{gray}2.06&\color{gray}2.33&12.52&13.61\\
        & Interleaved & \color{gray}2.57&\color{gray}2.92&\color{gray}2.30&\color{gray}2.71&\color{gray}2.17&\color{gray}2.45&11.48&12.52\\
    \cline{2-10}
        & Labeled & \color{gray}2.51&\color{gray}2.88&\color{gray}2.27&\color{gray}2.68&\color{gray}2.08&\color{gray}2.36&12.06&13.11\\
        & Multi-task Labeled & \color{gray}2.55&\color{gray}2.91&\color{gray}2.29&\color{gray}2.74&\color{gray}2.15&\color{gray}2.43&11.45&12.59\\
        & Weighted & \color{gray}2.21&\color{gray}2.44&\color{gray}2.13&\color{gray}2.41&\color{gray}2.04&\color{gray}2.18&\bf17.65&\bf18.31\\
    \hline\hline
        \multirow{6}{*}{GPT-2} & PersonaChat (single)&
        \color{gray}2.54&\color{gray}2.79&\color{gray}2.28&\color{gray}2.59&\color{gray}2.12&\color{gray}2.34&12.85&13.74\\
        & Concatenated & \color{gray}2.58&\color{gray}2.99&\color{gray}2.41&\color{gray}2.79&\color{gray}2.16&\color{gray}2.39&12.08&12.99\\
        & Interleaved & \color{gray}2.64&\color{gray}2.89&\color{gray}2.37&\color{gray}2.65&\color{gray}2.23&\color{gray}2.42&12.13&12.87\\
        \cline{2-10}
        & Labeled & \color{gray}2.57&\color{gray}2.84&\color{gray}2.32&\color{gray}2.62&\color{gray}2.16&\color{gray}2.37&12.37&13.10\\
        & Multi-task Labeled & \color{gray}2.65&\color{gray}2.90&\color{gray}2.37&\color{gray}2.65&\color{gray}2.22&\color{gray}2.42&12.14&12.86\\
        & Weighted & \color{gray}2.39&\color{gray}2.63&\color{gray}2.27&\color{gray}2.52&\color{gray}2.02&\color{gray}2.17&\bf15.96&\bf17.07\\

    \end{tabular}}
    \caption{$\alpha$DF$_d$ scores for generated responses from PersonaChat}
    \label{table:df-personachat}
\end{subtable}
\caption{Full results of $\alpha$DF$_d$ scores for generated responses from multiple corpora}
\end{table}

\bigskip


\section{Example of human evaluation system}\label{app:human_evaluation_system}

\begin{figure}[!ht]
    \centering
    \includegraphics[width=\textwidth]{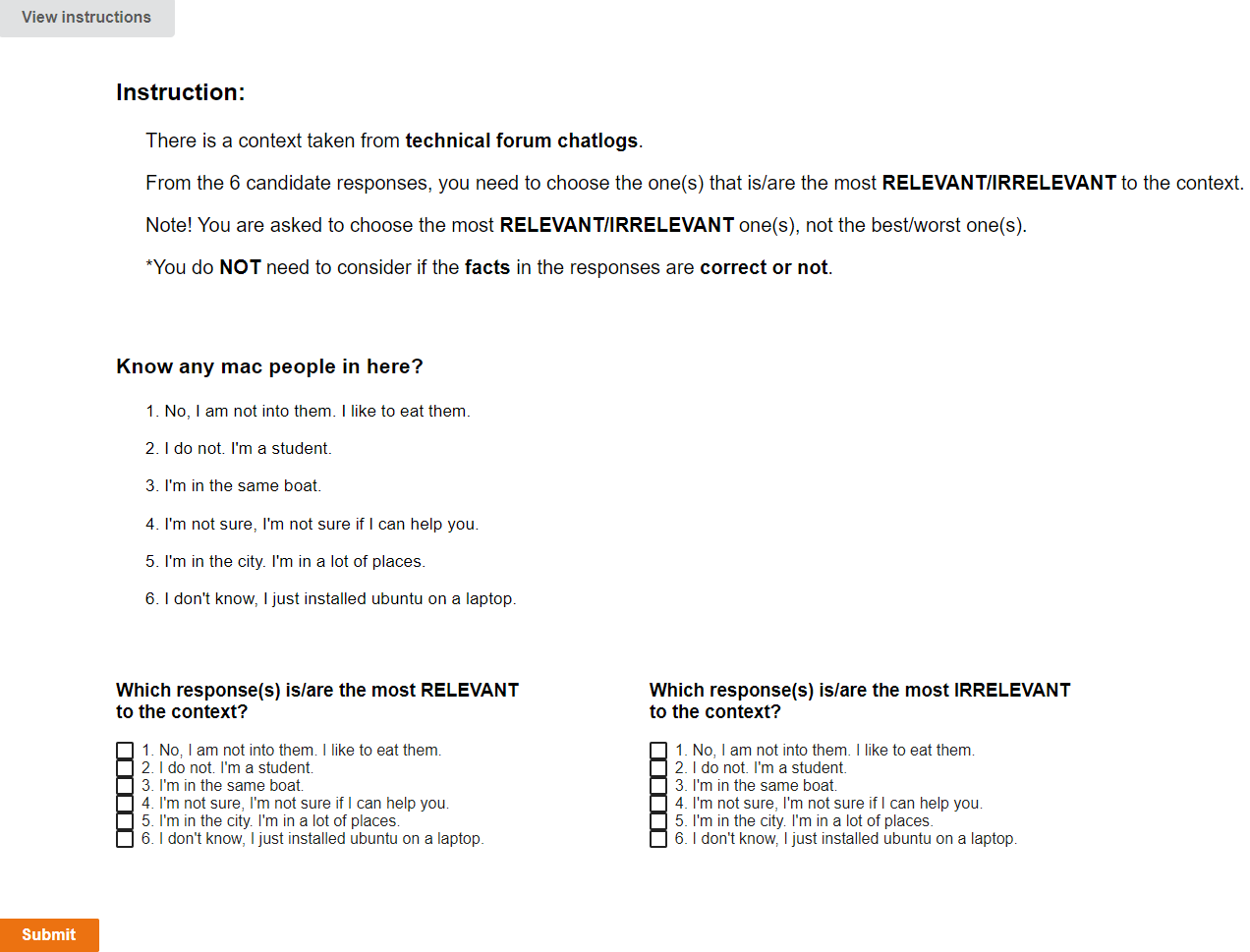}
    \caption{Human evaluation system for Ubuntu contexts}
    \end{figure}

\bigskip
\section{Examples of generated responses}\label{app:response-example}

\newpage

\begin{table*}[!ht]
\begin{minipage}{0.51\textwidth}
\centering
{\small
    \begin{tabular}{p{0.96\textwidth}}
    \toprule
        \cellcolor{r}\textbf{OSDB Context:} You know the main reason parents kill kids?\\
    \hline\hline
        \textbf{PersonaChat (single):} I am not sure. I am a vegan.\\
        \textbf{Concatenated:} I know. I've a daughter that just turned 18.\\
        \hline
        \textbf{Interleaved:} Because they're afraid of what they're gonna do.\\
        \textbf{Labeled:} Because they're not going to be able to handle it.\\
        \textbf{Multi-Task Labeled:} Because they're afraid of what they're gonna do.\\
        \textbf{Weighted:} Because they're afraid of the consequences.\\
    \toprule
    
    \cellcolor{r}\textbf{OSDB Context:} The big banks will fail, triggering a domino effect.\\
    \hline\hline
        \textbf{PersonaChat (single):} I am not sure yet. I am sure they will fail.\\
        \textbf{Concatenated:} I'm not sure. I'm not sure what the big banks are doing.\\
        \textbf{Interleaved:} What?\\
        \hline
        \textbf{Labeled:} I'm not sure what you're talking about.\\
        \textbf{Multi-Task Labeled:}I'm not sure what you mean.\\
        \textbf{Weighted:} What's going on?\\
    \toprule
    
    \cellcolor{r}\textbf{OSDB Context:} Rigor mortis has yet to set in. His core body temperature remains at 98 degrees. His skin color, lividity is healthy.\\
    \hline\hline
        \textbf{PersonaChat (single):}I'm not sure if I like him or not. I like to eat meat.\\
        \textbf{Concatenated:}That is good. I like to cook. I am a chef.\\
        \textbf{Interleaved:} He is a very good doctor.\\
        \hline
        \textbf{Labeled:} He's a good guy.\\
        \textbf{Multi-Task Labeled:} He is a man of great character.\\
        \textbf{Weighted:} He's been in the hospital for two weeks.\\
    \bottomrule\\\toprule

        \cellcolor{y}\textbf{Twitter Context:} Hillary ran away from rain today. Trump is back on stage minutes after assassination attempt.\\
    \hline\hline
        \textbf{PersonaChat (single):} I'm not sure what that means.\\
        \textbf{Concatenated:} That is so sad. I hope she gets better soon.\\
        \textbf{Interleaved:} I'm so sorry. I'm so sorry.\\
        \hline
        \textbf{Labeled:} I'm so sorry.\\
        \textbf{Multi-Task Labeled:} I'm not sure what happened to her.\\
        \textbf{Weighted:} She's a liar.\\
    \toprule
    
    \cellcolor{y}\textbf{Twitter Context:} It was gruesome \& extremely rare to have two witnesses. It was linked to epstein. Media didn't report it.\\
    \hline\hline
        \textbf{PersonaChat (single):} I'm sorry to hear that. I'm a nurse.\\
        \textbf{Concatenated:} I'm not sure what that means. I'm a big fan of the show.\\
        \textbf{Interleaved:} I'm not sure what happened to the epstein case.\\
        \hline
        \textbf{Labeled:} I'm not sure what the link was, but it was a link to a video.\\
        \textbf{Multi-Task Labeled:} I'm not sure what happened to the epstein case.\\
        \textbf{Weighted:} The media is biased and biased.\\
    \toprule
    
    \cellcolor{y}\textbf{Twitter Context:} This is 100\% true, but can we talk about it tomorrow and make argentina the focus today? Something needs to stick.\\
    \hline\hline
        \textbf{PersonaChat (single):} Yes, I think so. I'm going to go to the gym tomorrow.\\
        \textbf{Concatenated:} I'm going to go see a movie.\\
        \textbf{Interleaved:} I'm not sure what that's. I'm not sure what that's.\\
        \hline
        \textbf{Labeled:} I'm not sure what that means. I'm not sure what that means.\\
        \textbf{Multi-Task Labeled:} I'm not sure what you mean by that. I'm not sure what you mean by that.\\
        \textbf{Weighted:} I'm not sure if I can talk about it tomorrow.\\
    \bottomrule
    
    \end{tabular}}
\end{minipage}\begin{minipage}[t]{0.01\linewidth}\quad\end{minipage}\begin{minipage}{0.47\textwidth}
{\small
    \begin{tabular}{p{0.97\textwidth}}
    \toprule
         \cellcolor{b}\textbf{Ubuntu Context:} What are you going to do on the remote system exactly?\\
    \hline\hline
        \textbf{PersonaChat (single):} I am going to be a pilot. I am going to fly planes.\\
        \textbf{Concatenated:} I am going to go to the beach.\\
        \hline
        \textbf{Interleaved:} I'm going to install the latest version of Ubuntu.\\
        \textbf{Labeled:} I'm going to install a new kernel.\\
        \textbf{Multi-Task Labeled:} I'm going to install the latest version of Ubuntu.\\
        \textbf{Weighted:} I'm trying to connect to the server.\\

    \toprule
    
         \cellcolor{b}\textbf{Ubuntu Context:} Know any mac people in here?\\
    \hline\hline
        \textbf{PersonaChat (single):} No, I am not into them. I like to eat them.\\
        \textbf{Concatenated:} I do not. I'm a student.\\
        \textbf{Interleaved:} I'm in the same boat.\\
        \hline
        \textbf{Labeled:} I'm not sure, I'm not sure if I can help you.\\
        \textbf{Multi-Task Labeled:} I'm in the city. I'm in a lot of places.\\
        \textbf{Weighted:} I don't know, I just installed ubuntu on a laptop.\\

    \toprule
         \cellcolor{b}\textbf{Ubuntu Context:} Type sudo name of application (in short)\\
    \hline\hline
        \textbf{PersonaChat (single):} I like to play games on my computer.\\
        \textbf{Concatenated:} I did not see that. I am a programmer.\\
        \textbf{Interleaved:} I have a root shell, but I can't login.\\
        \hline
        \textbf{Labeled:} I have a root shell.\\
        \textbf{Multi-Task Labeled:} I don't know what that is.\\
        \textbf{Weighted:} It is Sudo apt-get install xubuntu-desktop.\\
    \bottomrule\\\toprule
         \cellcolor{p}\textbf{PersonaChat Context:} That must be hard. I like a beer after driving my nissan pathfinder.\\
    \hline\hline
        \textbf{PersonaChat (single):} I like to go to the movies. I have been to a few.\\
        \textbf{Concatenated:} I am a huge gamer, I play dungeons and dragons on my computer.\\
        \textbf{Interleaved:} I like to drive my nissan camry.\\
        \hline
         \textbf{Labeled:} I like to go to the movies with my friends.\\
         \textbf{Multi-Task Labeled:} I like to drive my prius.\\
        \textbf{Weighted:} I love beer too. I love to drink it while I am driving.\\
        
    \toprule
         \cellcolor{p}\textbf{PersonaChat Context:} A food bank near my house. Do you ever volunteer?\\
    \hline\hline
        \textbf{PersonaChat (single):} I do, I love to volunteer at the soup kitchen.\\
        \textbf{Concatenated:} Yes I do. I love to help people.\\
        \textbf{Interleaved:} I do. I have a lot of friends.\\
        \hline
         \textbf{Labeled:} I do. I am a volunteer at the local soup kitchen.\\
         \textbf{Multi-Task Labeled:} I do. I love it. I volunteer at the animal shelter.\\
        \textbf{Weighted:} I do. I love volunteering at the food pantry.\\
        
    \toprule
         \cellcolor{p}\textbf{PersonaChat Context:} Yes! My son is in junior high and I just started letting him watch them too.\\
    \hline\hline
        \textbf{PersonaChat (single):} I am a mom of two boys. I have a son.\\
        \textbf{Concatenated:} I am a mom, I have three kids.\\
        \textbf{Interleaved:} I love kids, I have a son who is in high school.\\
        \hline
         \textbf{Labeled:} I bet he is a good kid. I am a teacher.\\
         \textbf{Multi-Task Labeled:} I am sorry to hear that. I am a teacher.\\
        \textbf{Weighted:} I bet you are a good mom.\\
    \bottomrule
    \end{tabular}}
\end{minipage}
    \caption{Responses generated from GPT-2 fine-tuned on 4 corpora with multiple methods}
\end{table*}

\newpage

\end{document}